%% file: main.tex
\documentclass[lettersize,journal]{IEEEtran}
\usepackage{amsmath,amsfonts}
\usepackage{amssymb}
\usepackage{algorithmic}
\usepackage{algorithm}
\usepackage{array}
\usepackage{textcomp}
\usepackage{stfloats}
\usepackage{url}
\usepackage{verbatim}
\usepackage{graphicx}
\usepackage{subcaption}
\usepackage{wrapfig}
\usepackage{multirow}
\usepackage{marvosym}
\usepackage{booktabs}
\usepackage{cite}
\usepackage{pifont}
\usepackage{xcolor}
\usepackage[table]{xcolor}
\usepackage[pagebackref,breaklinks,colorlinks,citecolor=blue]{hyperref}
\definecolor{best}{rgb}{1.0, 0.6, 0}
\definecolor{best2}{rgb}{1.0, 0.8, 0.6}
\definecolor{bestlora}{rgb}{1.0, 0.8, 0.6}

\newcommand{\insertfig}{\includegraphics[width=0.92\linewidth]{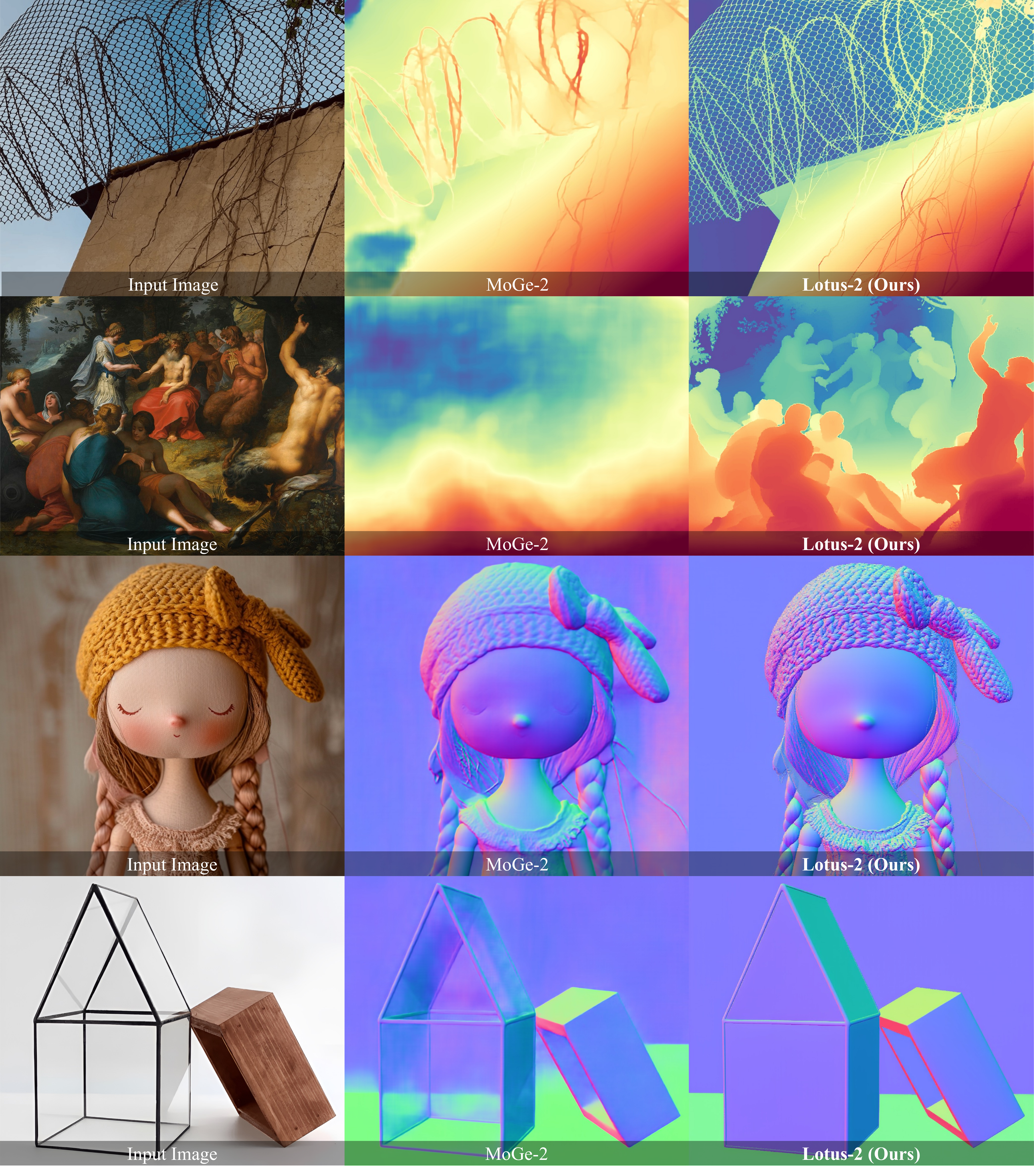}\captionof{figure}{
    \textbf{We present Lotus-2, a two-stage deterministic framework for monocular geometric dense prediction.} 
    Our method leverages pre-trained generative model as a deterministic world prior to achieve \textbf{new state-of-the-art accuracy} while requiring \textbf{remarkably minimal data} (trained on only $0.66\%$ of the samples used by MoGe-2~\cite{wang2025moge2}). The decoupled, two-stage design ensures both \emph{structurally correct} inference and \emph{high-fidelity} detail refinement. This figure demonstrates Lotus-2's robust zero-shot generalization with sharp geometric details, especially in challenging cases like oil paintings and transparent objects.
}
\label{fig:teaser}
}
\makeatletter
\apptocmd{\@maketitle}{\centering\setcounter{figure}{0}\vspace{1.2em}\insertfig\vspace{-1em}}{}{}
\makeatother

\begin{document}

\title{Lotus-2: Advancing Geometric Dense Prediction with Powerful Image Generative Model}

\author{
Jing He, 
Haodong Li\textsuperscript{*}, 
Mingzhi Sheng\textsuperscript{*}, 
Ying-Cong Chen\textsuperscript{\Letter}
% IEEE Publication Technology,~\IEEEmembership{Staff,~IEEE,}
        % <-this % stops a space
\thanks{Work done at the Hong Kong University of Science and Technology
  (Guangzhou). * denotes equal contribution. The authors' e-mail addresses are: $\{$jhe812, hli736, msheng758$\}$@connect.hkust-gz.edu.cn, yingcongchen@ust.hk. Corresponding Author: Ying-Cong Chen.}% <-this % stops a space
}

\markboth{}
{Shell \MakeLowercase{\textit{et al.}}: Lotus-2: Advancing Geometric Dense Prediction with Powerful Image Generative Model}


\maketitle

\begin{abstract}
Recovering pixel-wise geometric properties from a single image is fundamentally ill-posed due to appearance ambiguity and non-injective mappings between 2D observations and 3D structures. While discriminative regression models achieve strong performance through large-scale supervision, their success is bounded by the scale, quality, and diversity of available data, as well as by limited physical reasoning. Recent diffusion models exhibit powerful \emph{world priors} that encode geometry and semantics learned from massive image–text data, yet directly reusing their stochastic generative formulation is suboptimal for deterministic geometric inference: the former is optimized for diverse and high-fidelity image generation, whereas the latter requires stable and accurate predictions. 
In this work, we propose Lotus-2, a two-stage deterministic framework for stable, accurate and fine-grained geometric dense prediction, aiming to provide an optimal adaptation protocol to fully exploit the pre-trained generative priors. 
Specifically, in the first stage, the core predictor employs a single-step deterministic formulation with a clean-data objective and a lightweight local continuity module (LCM) to generate globally coherent structures without grid artifacts. 
In the second stage, the detail sharpener performs a constrained multi-step rectified-flow refinement within the manifold defined by the core predictor, enhancing fine-grained geometry through noise-free deterministic flow matching.
Using only 59K training samples—less than 1\% of existing large-scale datasets—Lotus-2 establishes new state-of-the-art results in monocular depth estimation and highly competitive surface normal prediction. These results demonstrate that diffusion models can serve as deterministic world priors, enabling high-quality geometric reasoning beyond traditional discriminative and generative paradigms. Project page: \href{https://lotus-2.github.io/}{\textcolor{blue}{\fontfamily{cmtt}\selectfont{lotus-2.github.io}}}. 
\end{abstract}
\begin{IEEEkeywords}
Dense prediction, depth estimation, normal estimation, diffusion model, rectified flow, image generation.
\end{IEEEkeywords}

\input{sec/0-introduction-yc}

\input{sec/1-rw_v2}
\input{sec/2-pre_v2}
\input{sec/3-lotus2_v3}
\input{sec/4-exp_v2}
\input{sec/5-conclusion_v2}

\input{sec/acknowledgments}
\input{sec/appendix}

% \newpage
\bibliographystyle{IEEEtran}
\bibliography{references}

% \newpage

% \input{sec/biography}

% \vfill
\end{document}

%% file: sec/0-introduction-yc.tex
\section{Introduction}
Geometric dense prediction aims to recover pixel-wise geometric or physical properties, such as depth, surface normal, or albedo, from a single image. 
% Geometric dense prediction from a single image aims to recover pixel-wise geometric or physical properties, such as depth, surface normal, or albedo, that are globally consistent with the underlying 3D scene. 
This problem lies at the foundation of modern visual understanding and serves as a cornerstone for various downstream applications, including controllable image generation~\cite{zhang2023adding, hu2024animate}, 3D/4D reconstruction~\cite{huang20242d, long2024wonder3d, jiang2025dimer}, and autonomous driving~\cite{li2023fb, li2024bevformer,gu2024dome}. 
The mapping from image appearance to underlying geometry is inherently ill-posed: a single image can correspond to multiple plausible 3D interpretations. 
% The task is inherently ill-posed: multiple distinct 3D structures can give rise to nearly identical image observations. 
Consequently, a model must infer a physically plausible and globally coherent structure beyond what is directly observable from appearance. 

Traditional approaches have long attempted to solve this problem through either geometric reasoning or discriminative learning. 
Early multi-view geometry and photometric consistency methods rely on strong assumptions about scene structure, lighting, and reflectance, making them unsuitable for single-view and complex real-world scenarios. 
With the rise of deep learning, discriminative models~\cite{eigen2014depth, yuan2022neural, eftekhar2021omnidata, yang2024depth, yang2024depth2, wang2025moge, wang2025moge2,li2024bi} have become the dominant paradigm by directly regressing geometric quantities from single images. 
While such models have achieved remarkable progress through increasingly powerful architectures and large-scale training, their performance remains fundamentally constrained by the scale, quality and diversity of available data. Human perception leverages strong world priors to resolve the ambiguity of geometric dense prediction, however, discriminative models trained on limited data distributions lack such mechanisms. 
% The mapping from image appearance to underlying geometry is inherently ill-posed: a single image can correspond to multiple plausible 3D interpretations. 
% While human perception leverages strong world priors to resolve such ambiguities, discriminative models trained on finite data distributions lack such mechanisms. 
Consequently, they perform poorly in rare and challenging scenes, involving transparency, reflection, and low texture, where inference requires reasoning beyond observable appearance. 
Even recent large-scale efforts---such as MoGe~\cite{wang2025moge, wang2025moge2} and DepthAnything~\cite{yang2024depth, yang2024depth2}, trained on millions of samples---still rely heavily on distributional coverage rather than true scene understanding from world modeling, see Fig. \ref{fig:teaser} for reference. 

The emergence of diffusion models such as Stable Diffusion~\cite{rombach2022high} and FLUX~\cite{bfl2024flux} has revealed a new paradigm for visual reasoning. 
Trained on billions of diverse image–text pairs (\emph{e.g.}, LAION-5B~\cite{schuhmann2022laion}), these models exhibit remarkable capability in synthesizing geometrically coherent and physically consistent imagery across diverse scenes. 
This success suggests that diffusion backbones implicitly encode \emph{world priors}—rich internal representations of geometry and semantics accumulated through large-scale generative training.

With this intuition, recent works have attempted to repurpose the world priors for dense prediction~\cite{lee2024exploiting, ke2024repurposing, he2024lotus, li20252, wang2025jasmine, fu2024geowizard, zhao2025diception}. 
While these studies validate the promise of generative world priors, most of them directly adopt the original generative formulation of diffusion models without rethinking its suitability for dense prediction. 
For example, Marigold~\cite{ke2024repurposing} fine-tunes Stable Diffusion by reformulating depth estimation as an image-conditioned depth generation problem. 
Although this design benefits from the pre-trained priors, it overlooks the fundamental difference between dense prediction and image generation: the former requires deterministic and accurate inference, whereas the latter optimizes for diverse and high-fidelity generation through stochastic multi-step sampling. 
% the former requires deterministic, geometrically constrained inference, whereas the latter optimizes for diverse generation through stochastic sampling. 
This fundamental mismatch often results in inconsistent and inaccurate geometric structure.
Post-processing (\textit{e.g.}, test-time ensembling~\cite{ke2024repurposing,fu2024geowizard}) doesn't solve it in a native manner, and needs repeated predictions and may produce blurry results.

Motivated by these limitations, we revisit the role of diffusion-based generative models in dense prediction and propose a new perspective: their true value lies not in the generative mechanism itself, but in the \emph{world priors} encoded within their pre-trained weights. 
% but in the \emph{world modeling capability} encoded within their pre-trained weights. 
Instead of treating diffusion as a stochastic generator, we view it as a structured world prior that can guide the inference towards deterministic and geometrically accurate dense prediction. 
% Instead of treating diffusion as a stochastic generator, we view it as a structured world prior that can guide deterministic inference toward geometrically and physically consistent predictions. 
Based on this insight, we introduce \emph{Lotus-2}, a two-stage deterministic framework that decouples accurate global geometry prediction from meticulous detail sculpting, effectively combining the strengths of regression and generative expressiveness.
% Based on this insight, we introduce \emph{Lotus-2}, a two-stage framework that decouples structure prediction from detail refinement, effectively combining the strengths of regression-based determinism and generative expressiveness. 

In the first stage, a \emph{core predictor} extracts globally coherent and accurate geometry through a simple yet effective adaptation of the rectified-flow formulation in FLUX~\cite{bfl2024flux}. 
By systematically analyzing the key designs of stochastic generative formulation, including the stochasticity, multi-step sampling and parameterization type, we identify that a single-step deterministic formulation under a clean-data prediction yields much more stable and accurate results than the original stochastic multi-step residual-based design.
% By systematically analyzing the effects of time-step count and parameterization type, we identify that a single-step formulation under a clean-data prediction target yields more stable and precise results than the original multi-step residual-based design. 
This single-step predictor is further enhanced with a lightweight \emph{local continuity module (LCM)}, which mitigates grid artifacts introduced by the non-parametric Pack–Unpack operations in FLUX while maintaining architectural compatibility and efficiency. 

In the second stage, an \textit{optional} \emph{detail sharpener} performs a detail refinement through a deterministic multi-step rectified-flow process. 
It operates within the constrained manifold defined by the core predictor and learns the transition from the ``accurate'' to ``accurate and fine-grained'' annotation, progressively enriching geometric details while preserving global structure and accuracy. 
% Rather than generating arbitrary variations, it operates within the local manifold defined by the core predictor, progressively enriching geometric details while preserving global structure and accuracy. 
This design bridges the gap between regression and generative modeling: the former ensures structural stability and correctness, while the latter contributes fine-grained realism. 
Consequently, Lotus-2 effectively leverages the generative priors in a disciplined and interpretable manner, achieving both geometric consistency and high-frequency detail fidelity without sacrificing efficiency and stability.

In summary, our key contributions are:

\begin{itemize}
    \item \textbf{Revisiting the role of diffusion models for dense prediction.} 
    We reformulate diffusion-based generative models from stochastic image generators to structured world priors, emphasizing that their strength lies in the world modeling capability embedded within pre-trained weights rather than in the sampling trajectory itself. 

    \item \textbf{A two-stage deterministic framework integrating the strengths of regression and generative refinement.}
    We propose \emph{Lotus-2}, which decouples structure prediction and detail refinement: 
    a \emph{core predictor} performs single-step, clean-data regression for accurate and stable geometric estimation, while an optional \emph{detail sharpener} applies multi-step rectified-flow refinement within the constrained manifold defined by the predictor. 

    \item \textbf{A principled adaptation of the rectified-flow formulation.}
    Through systematic analysis of several key designs in the original stochastic generative formulation, including stochasticity, multi-step sampling, parameterization type and local continuity, we demonstrate that the single-step clean-data deterministic design achieves higher accuracy and better optimization stability than the traditional formulation optimized for image generation. 
    % Furthermore, a lightweight \emph{local continuity module (LCM)} effectively suppresses grid artifacts without sacrificing efficiency. 

    \item \textbf{State-of-the-art performance.}
    With only 59K training samples—merely $0.66\%$ of the data used by MoGe~\cite{wang2025moge, wang2025moge2} and $0.09\%$ of that used by DepthAnything~\cite{yang2024depth, yang2024depth2}, Lotus-2 achieves new state-of-the-art results on monocular depth estimation and highly competitive results on normal estimation.
    % It delivers high-fidelity details and geometric consistency while requiring at most 11 inference steps, significantly faster than existing generative counterparts such as Marigold~\cite{ke2024repurposing} and GeoWizard~\cite{fu2024geowizard}.
\end{itemize}

%% file: sec/1-rw_v2.tex
\section{Related Work}
\subsection{Traditional Paradigms for Geometric Dense Prediction}
Recovering geometric properties, specifically monocular depth estimation and surface normal prediction, has been a central pursuit in computer vision. Traditional efforts to address the inherent ill-posed nature of this task have generally been categorized into two major paradigms: (1) physics-based geometric reasoning and (2) data-driven discriminative learning. 

Early physics-based geometric reasoning methods focus on leveraging established geometric and photometric constraints, such as structure from motion (SfM)~\cite{tomasi1992shape,snavely2008modeling}, photometric stereo~\cite{woodham1980photometric}, and algorithms based on multi-view geometry~\cite{scharstein2002taxonomy, hartley2003multiple}. They rely on a set of strong assumptions about the scene. For instance, they often require multiple views of the scene, precise camera calibration, or strict adherence to the Lambertian reflectance model. While theoretically sound under constrained or ideal conditions, these dependencies render them brittle and highly impractical for single-view geometric dense prediction in unconstrained, real-world environments. 

With the advent of deep learning, the field shifts toward the discriminative learning paradigm. Models like the pioneering works~\cite{eigen2014depth, eftekhar2021omnidata, ranftl2020towards, ranftl2021vision} and the recent large-scale state-of-the-art efforts, such as MoGe~\cite{wang2025moge, wang2025moge2} and Depth Anything~\cite{yang2024depth, yang2024depth2}, have achieved remarkable empirical performance by directly regressing geometric quantities from input images. These successes are primarily attributed to increasingly powerful architectures (like Vision Transformers~\cite{dosovitskiy2020image}) and, more importantly, supervision at massive scales. 

However, these discriminative models face two fundamental limitations rooted in their data-driven nature. First, despite the scale of modern datasets, the quantity, quality, and diversity of geometric ground truth data remain fundamentally constrained (only million-scale compared to the billion-scale data used for pre-training large-scale generative models). Second, these models rely heavily on distributional coverage—memorizing patterns across the training data—rather than learning the intrinsic physical laws that govern scene structure. Consequently, they struggle severely with out-of-distribution (OOD) scenarios, including highly reflective surfaces, transparent objects, or rare scene compositions, where inference requires true geometric reasoning that transcends memorized patterns. These limitations motivate us to explore the powerful world priors embedded within pre-trained large-scale generative models. The world priors offer a superior foundation because: 
(1) they have been exposed to vast amounts of high-quality data; and (2) they possess world intrinsic knowledge of geometry, semantics, and physical structure accumulated through large-scale generative training. 

\subsection{World Priors from Generative Models}
The dependence of data-driven discriminative models on finite supervised data underscores the necessity of a superior source of structural knowledge, divorced from expensive geometric annotation. This required ``world prior'' has been implicitly encoded within the weights of large-scale generative models. 

Unlike early generative approaches such as VAEs~\cite{van2017neural,razavi2019generating} and GANs~\cite{goodfellow2014generative, pixelfolder, StyleGAN1, StyleGAN2, StyleGAN3}, and even initial diffusion models~\cite{ho2020denoising, song2020denoising,li2024discene,liang2024luciddreamer,yang2025advancing,he2024disenvisioner}, which are often trained on restricted domain-specific data and thus contain limited world knowledge, recent advancements focus on large-scale training. By leveraging billions of diverse image-text pairs (\emph{e.g.}, LAION-5B~\cite{schuhmann2022laion}), modern diffusion models have acquired an extraordinary capacity to synthesize geometrically coherent and physically consistent imagery across diverse scenes. Crucially, the sheer volume of the training data significantly surpasses the quantity of all available dense geometric annotation datasets~\cite{wang2020tartanair,li2018megadepth,wang2019irs,cho2021diml,yao2020blendedmvs}. This success implies that the diffusion backbone implicitly encodes powerful world priors—rich internal representations of geometry, semantics, and physical structure—thereby offering a new paradigm for visual reasoning. 

The landscape of these pre-trained large-scale diffusion models has rapidly evolved: StabilityAI's release of Stable Diffusion 1.x and 2.x~\cite{rombach2022high,podell2023sdxl}, based on the DDPM~\cite{ho2020denoising} training paradigm and a UNet~\cite{ronneberger2015u} structure, initially revolutionizes the field. Subsequent efforts focused on efficiency and quality, such as Playground's aesthetic enhancement efforts~\cite{li2024playground} and PixArt-$\alpha$'s exploration~\cite{chen2023pixart} of the DiT~\cite{peebles2023scalable} structure for computational efficiency. More recently, the emergence of the rectified-flow~\cite{liu2022flow} and flow-matching~\cite{lipman2022flow} formulations, explored in models like Stable Diffusion 3.x~\cite{esser2024scaling}, AuraFlow~\cite{cloneofsimo2024auraflow}, and significantly, FLUX~\cite{bfl2024flux}, represents the latest technological frontier. FLUX built upon the DiT architecture and the rectified-flow formulation, achieves the highest aesthetic quality through meticulous training and data preparation, leading to exceptionally natural, realistic, and geometrically consistent visual synthesis. Given the visual quality and superior physical consistency, the pre-trained FLUX model is the optimal choice as the world prior for our geometric dense prediction. 

\subsection{Repurposing Generative Priors for Dense Prediction}
Building upon the insight that pre-trained generative weights encode crucial world priors, the community has explored various strategies to adapt this knowledge for geometric dense prediction. These methods can be broadly categorized into three distinct technical trajectories. The most dominant group follows the ``stochastic generative formulation'', retaining the original multi-step diffusion pipeline. This includes works like Marigold~\cite{ke2024repurposing}, GeoWizard~\cite{fu2024geowizard}, DepthFM~\cite{gui2025depthfm} and recent DICEPTION~\cite{zhao2025diception}. While these models validate the necessity of world knowledge from generative priors, their adherence to ``stochastic multi-step sampling'' leads to fundamental performance limitations: poor inference efficiency and unacceptable structural variance due to their non-deterministic nature. All of these methods rely on random noise, and different noises result in diverse geometric structure. This diversity is desirable in image generation, however, it results in inconsistent and physically implausible geometric structures in dense geometric prediction. A second group focuses on accelerating the inference speed. Works like Diffusion-E2E-FT~\cite{garcia2025fine, he2024lotus} directly fine-tune the generative backbone as a deterministic feed-forward model to achieve fast, stable results. However, this single-step strategy often struggles to produce the fine-grained geometric details, which is crucial for high fidelity. The third group attempts a coarse-to-fine strategy, exemplified by StableNormal~\cite{ye2024stablenormal}, which uses a two-stage approach combining initial prediction with subsequent refinement. However, its second stage still relies on the stochastic generative formulation, compromising the inherent need for high stability in geometric inference. In contrast to all these approaches, our proposed Lotus-2 employs  a  purely deterministic and noise-free rectified-flow strategy for both stages of prediction. By utilizing the superior FLUX backbone and decoupling the inference into structure prediction (core predictor) and detail refinement (detail sharpener), we overcome the limitations of stochasticity, efficiency, and detail loss, positioning Lotus-2 as the premier solution for physically consistent and fine-grained geometric reasoning. 

%% file: sec/2-pre_v2.tex
\section{Preliminaries}
Our Lotus-2 framework is founded on the mathematical formalism of rectified-flow and the architectural foundation of FLUX model. This section introduces the necessary technical background related to our methodology. 

\subsection{Rectified-Flow Formulation}
\label{sssec:pre-rf}
The rectified-flow (RF) formulation~\cite{liu2022flow,lipman2022flow} provides a robust and deterministic framework for modeling the transformation between two arbitrary probability measures via an ordinary differential equation (ODE). 
Specifically, given a source distribution $p_1$ and a target distribution $p_0$, the ODE on time-step $ t\in[0,1]$ is defined as: $d\mathbf{z_t} = v(\mathbf{z_t}, t) dt$, which maps $\mathbf{z_1} \sim p_1$ to $\mathbf{z_0} \sim p_0$ under the velocity vector field $v(\mathbf{z_t}, t)$. Crucially, the core principle of RF is to transport samples along the straight-line path: 
\begin{equation}
    \mathbf{z_t} = t\mathbf{z_1}+(1-t)\mathbf{z_0},   
\end{equation}
thus the target vector field $\mathbf{v}$ is given by $\mathbf{v}=\frac{d\mathbf{z_t}}{dt}=\mathbf{z_1}-\mathbf{z_0}$. 
This straight-line mechanism fundamentally differs from the high-curvature paths of denoising diffusion models~\cite{ho2020denoising}, which ensures a high efficiency and reduced error accumulation. 
For training, the velocity vector field $v(\mathbf{z_t}, t)$ is parameterized by a neural network $f_\theta$, which is optimized by minimizing the distance to the target vector field $\mathbf{v}$. The loss function is thus defined as: 
\begin{align}
\label{eq:base-rf}
    L_t &= {||\mathbf{v} - f_\theta(\mathbf{z_t}, t)||}^2 \\
        &= {||(\mathbf{z_1}-\mathbf{z_0}) - f_\theta(\mathbf{z_t}, t)||}^2.
\end{align}
In practice, the expectation over the continuous time $t\in[0,1]$ is approximated by randomly sampling a discrete time-step value from a pre-defined set at each training iteration. Given a total of $T$ training time-steps, the pre-defined  time-step set is:
\begin{equation}
\{ t_i = \frac{i}{T} \mid i = 1, 2, \dots, T \}.
\label{eq:time-set}
\end{equation}
During sampling (inference), the discrete Euler solver is used to iteratively generate the target sample ($t=0$) from the source ($t=1$). Formally, the  iterative sampling process from current state $\mathbf{z_{t_{curr}}}$ to next state $\mathbf{z_{t_{next}}}$ is given by: 
\begin{equation}
\label{eq:euler}
    \mathbf{z_{t_{next}}} = \mathbf{z_{t_{curr}}} - \eta\cdot f_\theta(\mathbf{z_{t_{curr}}},t), 
\end{equation}
where $t_{\text{next}}< t_{\text{curr}}$ and $\eta$ ($0<\eta\leq1$) denotes the step size, which is determined by the total number of inference time-steps $T_{\text{inf}}$. 
% the step size $\mathbf{\eta_t}$ is calculated as: 
% \begin{equation}
    % \mathbf{\eta_t}=\mathcal{S}_\mathbf{t}(\frac{1}{T_{\text{inf}}}). 
% \end{equation}

\subsection{Architectural Foundation of FLUX}
We leverage the architecture and weights of FLUX\cite{bfl2024flux}, which utilizes a pre-trained variational autoencoder (VAE) to compress high-dimensional image data $\textbf{x}$ into a compact latent space $\mathcal{Z}$. The VAE consists of an encoder $E$ and a decoder $D$, where $E(\textbf{x}) = \mathbf{z^x}$ maps the image to a latent code, and $D(\mathbf{z^x}) = \hat{\textbf{x}}$ attempts to reconstruct the image from the latent code. The rectified-flow formulation of FLUX operates within this VAE latent space $\mathcal{Z}$.

In the specific task of image generation, the starting distribution $p_1$ is set to standard Gaussian noise in the latent space, \emph{i.e.}, $\mathbf{z_1} \sim \mathcal{N}(0, I)$. The target distribution $p_0$ is the distribution of real, clean image latent, \emph{i.e.}, $\mathbf{z_0} =E(\textbf{x})= \mathbf{z^x}$. Based on this setup, the loss function in Eq.~\ref{eq:base-rf} is rewritten by: 
\begin{equation}
    L_t= {||(\mathbf{\epsilon} - \mathbf{z^x}) - f_\theta(\mathbf{z_t}, t)||}^2.
\end{equation}
Here, $\mathbf{z_t} = t\epsilon + (1-t)\mathbf{z^x}$ is the linear interpolation between the noise and the target latent code. FLUX adopts the DiT (Diffusion Transformer)~\cite{peebles2023scalable} architecture as its model $f_\theta$. 

\noindent \textbf{The Pack-Unpack Operations in FLUX. }
\label{sssec:pre-pack}
To reduce computational overhead and memory usage, FLUX applies paired \emph{Pack} and \emph{Unpack} operations around the DiT model in the latent space. Pack is a parameter-free down-sampling procedure that rearranges the latent feature by grouping every non-overlapping $2\times 2$ patch into the channel dimension, 
\begin{equation}
    \text{Pack}: \mathbb{R}^{H \times W \times C} \rightarrow \mathbb{R}^{\frac{H}{2} \times \frac{W}{2} \times 4C}.
\end{equation}
Conversely, Unpack restores the original resolution by inverting this rearrangement, 
\begin{equation}
    \text{Unpack} : \mathbb{R}^{\tfrac{H}{2} \times \tfrac{W}{2} \times 4C} \;\;\to\;\; \mathbb{R}^{H \times W \times C}.
\end{equation}
This Pack-Unpack operation, while efficient, introduces a critical challenge: because it is parameter-free, it can introduce noticeable local pixel discontinuities (``grid-artifacts''). This issue is especially severe under the single-step formulation, degrading the overall quality and realism of the outputs. 

%% file: sec/3-lotus2_v3.tex
\begin{figure*}
    \centering
    \includegraphics[width = 1.0\linewidth]{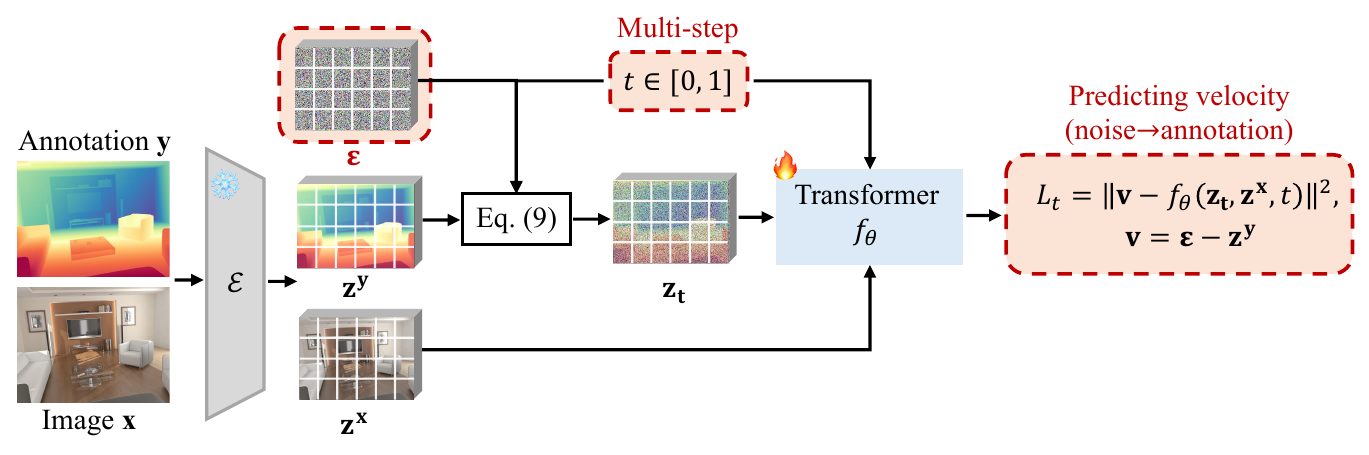}
    \caption{\textbf{Adaptation protocol of stochastic formulation (Stochastic-DA). } This framework models a conditional generative flow by estimating the velocity field from a random noise latent $\mathbf{\epsilon}$ to the annotation latent $\mathbf{z^y}$, conditioned on the image latent $\mathbf{z^x}$. The target velocity vector is $\mathbf{v} = \mathbf{\epsilon}-\mathbf{z^y}$. This reliance on noise initialization inherently leads to non-deterministic variance in deterministic geometric prediction. }
    \label{fig:sda}
\end{figure*}
\begin{figure*}
    \centering
    \includegraphics[width = 1.0\linewidth]{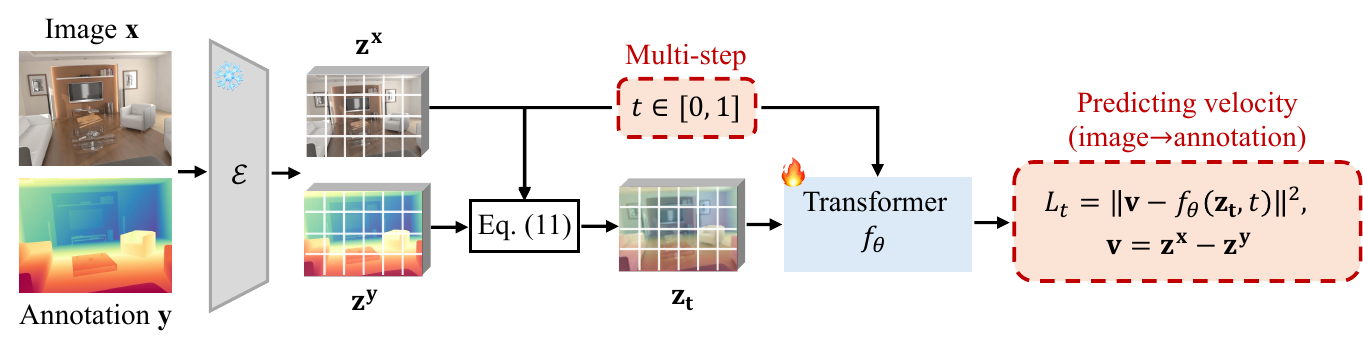}
    
    \caption{\textbf{Adaptation protocol of deterministic formulation (Deterministic-DA).} This architecture shifts the paradigm to a noise-free rectified-flow formulation. It directly estimates the velocity field from the source image latent $\mathbf{z^x}$ to the target annotation latent $\mathbf{z^y}$, where the target velocity vector is $\mathbf{v} = \mathbf{z^x}-\mathbf{z^y}$. This deterministic setup ensures stability and structural consistency for geometric dense prediction.
    % During training, the pre-trained VAE is kept fixed, and $f_\theta$ is fine-tuned with the LoRA strategy. To minimize modifications to the pre-trained model, we set the number of training time-steps to $T=1000$ and the number of inference time-steps to $T_{\text{inf}}=50$ as in FLUX.
    }
    \label{fig:dda}
\end{figure*}
\section{Lotus-2}
In this section, we present Lotus-2, a two-stage deterministic framework for stable, accurate and high-fidelity dense prediction, aiming to provide an optimal adaptation protocol to effectively and efficiently leverage the pre-trained world priors of FLUX~\cite{bfl2024flux}. 
% In this section, we present Lotus-2, a two-stage deterministic framework that re-purposes the generative world priors of FLUX~\cite{bfl2024flux} for stable, accurate and high-fidelity dense prediction. 
\textbf{We argue that directly inheriting the stochastic generative formulation---which is optimized for image synthesis---introduces instability and unnecessary complexity for deterministic geometric tasks.} The image synthesis aims at diverse and high-fidelity generation through stochastic multi-step sampling, while the dense prediction requires a deterministic and accurate inference. This fundamental misalignment results in high structural variance and significant prediction errors for dense prediction, thereby compromising overall accuracy. 
To better exploit the generative world priors, we propose a decoupled, two-stage adaptation protocol. 
We first introduce the \emph{Core Predictor} (Sec.~\ref{ssec:flux-lotus}) derived through a systematic analysis of the standard generative formulation, including its stochasticity (Sec.~\ref{sssec:s_vs_d}), multi-step iterative sampling (Sec.~\ref{sssec:num-timestep}), parameterization type (Sec.~\ref{sssec:param}), and local continuity (Sec.~\ref{sssec:lcm}). This core predictor is dedicated solely to achieving highly accurate and robust global geometry estimation.
% structural correctness. 
Subsequently, we address the challenge of fine-grained fidelity by proposing the \emph{Detail Sharpener} (Sec.~\ref{ssec:ds}), which employs a constrained multi-step rectified-flow formulation designed only for meticulous detail sculpting
% high-fidelity refinement 
within the established structural manifold. This decoupled, two-stage approach successfully achieves both structural accuracy and fine-grained fidelity, with its complete inference process detailed in Sec.~\ref{ssec:inference}. 

% In this section, we present Lotus-2, a two-stage deterministic framework that re-purposes the generative world prior of FLUX for geometrically consistent and physically coherent dense prediction. 
% We first establish our core design philosophy by contrasting stochastic and deterministic adaptation (Sec.~\ref{sssec:s_vs_d}). We then systematically analyze the inherited factors from the original rectified-flow formulation—the number of time-steps and parameterization type—to derive our highly efficient and accurate \emph{Core Predictor} (Sec.~\ref{sssec:num-timestep}, \ref{sssec:param}). We incorporate the \emph{Local Continuity Module} (LCM) into the core predictor to resolve ``grid artifacts'' caused by Pack-Unpack operations (Sec.~\ref{sssec:lcm}). Finally, we present the \emph{Detail Sharpener} for high-fidelity refinement (Sec.~\ref{ssec:ds}) and introduce the deterministic inference pipeline of our Lotus-2 (Sec.~\ref{ssec:inference}). 

\subsection{Core Predictor: Robust and Accurate Geometric Prediction}
\label{ssec:flux-lotus}
% In this section, we systematically analyze the key designs of stochastic generative formulation, including the stochasticity, multi-step sampling, parameterization type and local continuity, for robust and accurate geometric prediction. 

\begin{figure}
    \centering
    \includegraphics[width=1.0\linewidth]{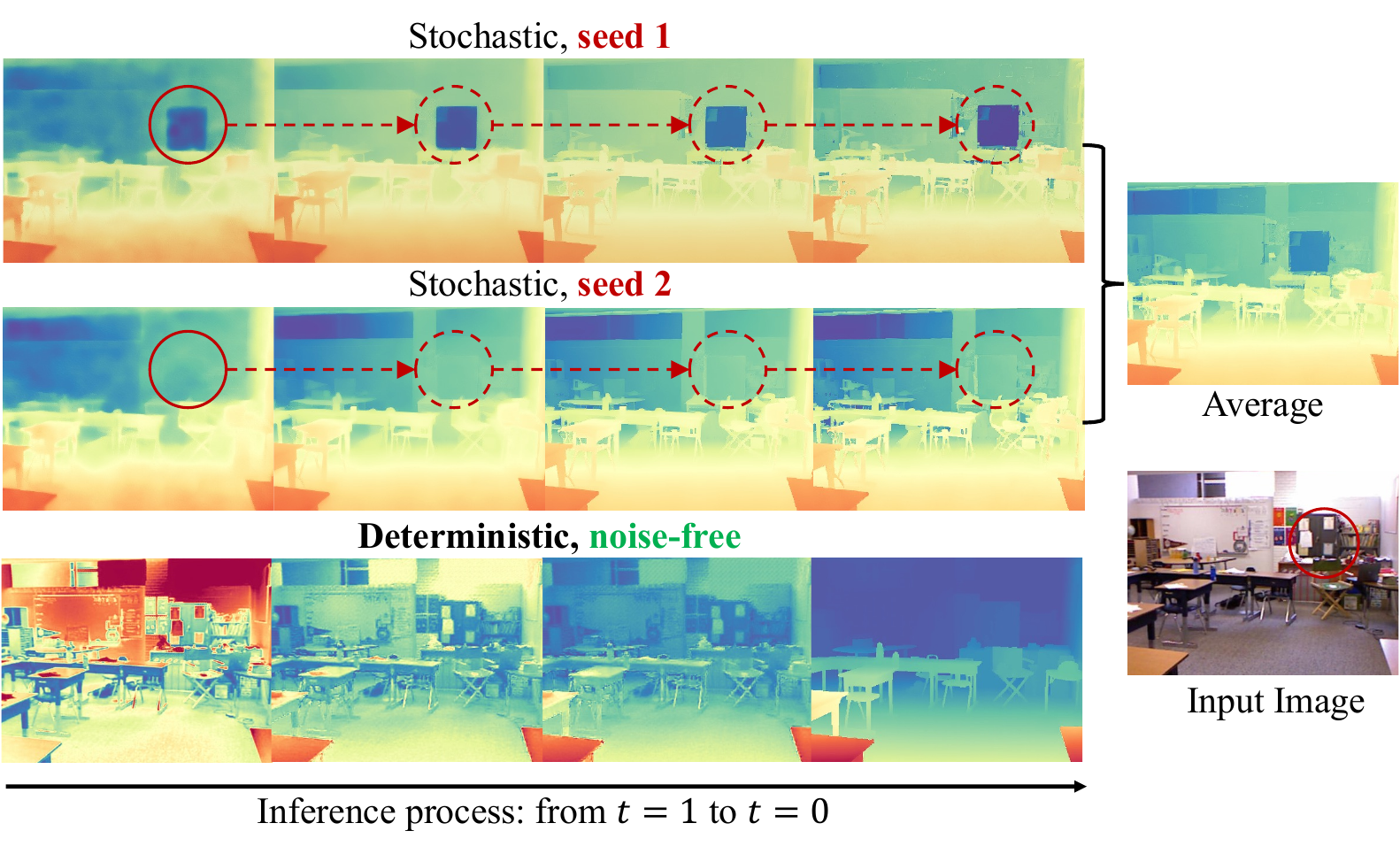}
    \caption{\textbf{Comparison between stochastic and deterministic formulation. } 
    % As detailed in Sec.~\ref{sssec:analysis-1}, we adapt the original multi-step rectified-flow for geometric dense prediction in stochastic and deterministic manners, respectively. 
    The figure visualizes the iterative inference process from $t=1$ to $t=0$. The stochastic formulation (Stochastic-DA) exhibits significant structural variance: distinct random noise initializations yield inconsistent geometric structures across the entire inference process (highlighted in \textcolor{red}{red circles}). While averaging is employed to mitigate the variance, the final prediction remains compromised by  the blending of conflicting structural hypotheses. In contrast, the deterministic formulation (Deterministic-DA) ensures a noise-free and stable trajectory, preventing structural variance and improving geometric coherence and prediction accuracy. }
    \label{fig:s_vs_d}
\end{figure}
\subsubsection{Analysis-1: Stochastic \emph{vs.} Deterministic Formulation}
\label{sssec:s_vs_d}
Initial efforts to leverage diffusion priors for geometric dense prediction (\emph{e.g.}, Marigold~\cite{ke2024repurposing}, GeoWizard~\cite{fu2024geowizard}) inherit the model's original \emph{stochastic generative formulation}. We term this approach as \emph{Stochastic Direct Adaptation} (Stochastic-DA). In this setup, the process is framed as an image-conditioned geometric generation task: the model learns the flow from pure Gaussian noise $\mathbf{\epsilon} \sim \mathcal{N}(0, I)$ to the target geometry $\mathbf{z^y}$ conditioned on the input image $\mathbf{z^x}$ as illustrated in Fig.~\ref{fig:sda}. 
Specifically, the latent feature at time $t$ is defined as: 
\begin{equation}
    \mathbf{z_t} = t\mathbf{\epsilon} + (1-t)\mathbf{z^y}. 
\end{equation}
The neural network $f_\theta$ is trained to predict the velocity field $\mathbf{v} = \mathbf{\epsilon} - \mathbf{z^y}$ by incorporating the image latent $\mathbf{z^x}$ as a conditional input (typically concatenated along the channel dimension of the input feature to the DiT backbone). The loss function for optimizing this stochastic generative formulation is given by: 
\begin{equation}
    L_t= {||(\mathbf{\epsilon} - \mathbf{z^y}) - f_\theta(\mathbf{z_t}, \mathbf{z^x}, t)||}^2.
\end{equation}

The core limitation of this approach is its inherent non-deterministic variance. Because the inference must begin with an initial sample of pure Gaussian noise, $\mathbf{z_1}=\epsilon\sim\mathcal{N}(0, I)$, different random initializations lead to diverse outputs, resulting in inconsistent geometric structures for the same input image, as illustrated in Fig. \ref{fig:s_vs_d}. This variance is beneficial for diverse image generation; however, it leads to physically implausible geometric structures for dense prediction, thus hindering accuracy. While ensemble averaging is commonly used to mitigate this variance, it inherently introduces prediction bias and also compromises overall accuracy by blending both correct and incorrect structural hypotheses.

% \noindent \textbf{The Deterministic Shift.}
To resolve this fundamental mismatch, we discard the stochastic conditional generative formulation and shift the paradigm to a purely deterministic flow matching between two distributions. We formulate the problem as learning a noise-free transformation between the image feature $\mathbf{z^x}$ and the geometric feature $\mathbf{z^y}$, directly utilizing the inherent determinism of the rectified-flow framework. We term this approach as \emph{Deterministic Direct Adaptation} (Deterministic-DA) of the rectified-flow formulation. The architecture for this approach is illustrated in Fig.~\ref{fig:dda}. 
Specifically, Deterministic-DA defines the two distributions as the image and annotation spaces, respectively: the source is the image latent $\mathbf{z^x}$ and the target is the annotation latent $\mathbf{z^y}$. The latent feature at time $t$ is defined as:
\begin{equation}
\mathbf{z_t} = t\mathbf{z^x} + (1-t)\mathbf{z^y},
\label{eq: direct-rf-forward-det}
\end{equation}
where the model $f_\theta$ is trained to predict the velocity $\mathbf{v}=\mathbf{z^x}-\mathbf{z^y}$. The training objective for this deterministic flow is: 
\begin{equation}
    L_t= {||(\mathbf{z^x}-\mathbf{z^y}) - f_\theta(\mathbf{z_t}, t)||}^2.
\end{equation}
This approach is inherently noise-free during both training and inference. As shown in Fig. \ref{fig:s_vs_d} and Tab. \ref{tab:ablation}, this deterministic approach significantly improves structural consistency and prediction accuracy compared to its stochastic counterpart. 

% \noindent \textbf{The Problem of Deterministic-DA.}
% While Deterministic-DA offers a clear performance benefit, it inherits the \emph{multi-step} formulation and the \emph{residual parameterization} from the original rectified-flow, which were optimized for high-fidelity image generation, not geometric precision. The subsequent analyses address these inherited limitations to derive our final core predictor design. 
\subsubsection{Analysis-2, Multi-Step Iterative Sampling} 
\label{sssec:num-timestep}
While the multi-step formulation enhances the capacity of generative models, it is optimized for high-fidelity image synthesis and demands large-scale training data. For dense geometric prediction, where high-quality supervision data is scarce, this inherited multi-step formulation is computationally intensive and makes the model difficult to optimize effectively. Furthermore, the prediction errors are accumulated during this multi-step iterative sampling, further compromising the accuracy. The iterative nature also hinders its practical application due to slow inference speeds. 

To address these challenges, we propose fine-tuning the pre-trained rectified-flow model with fewer training time-steps. 
\begin{figure*}
    \centering
    \includegraphics[width=0.95\linewidth]{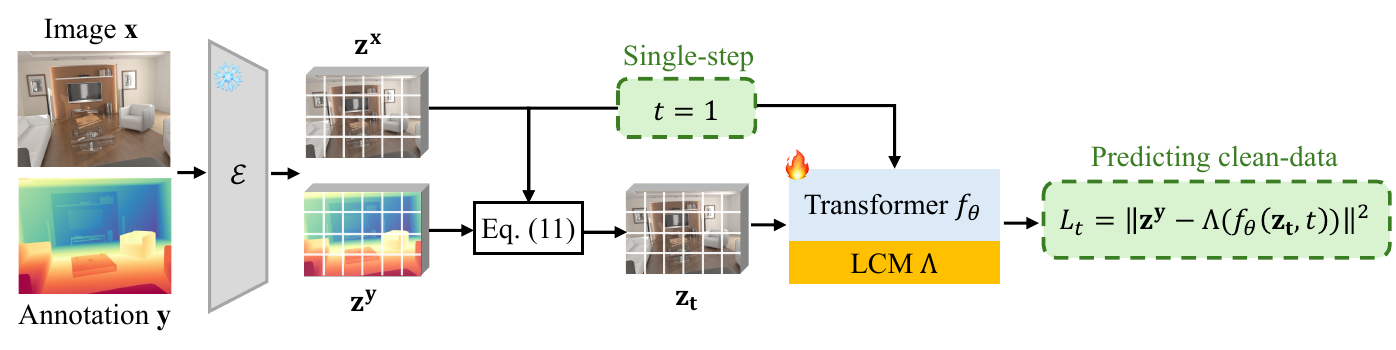}
    \caption{\textbf{Adaptation protocol of the core predictor in Lotus-2.} It adopts a single-step formulation ($t=1$) with clean-data prediction to efficiently exploit the world priors of pre-trained FLUX model, where input latent $\mathbf{z_t}$ is equivalent to the image latent $\mathbf{z^x}$, \emph{i.e}, $\mathbf{z_t}=\mathbf{z_1}=\mathbf{z^x}$, according to Eq.~\ref{eq: direct-rf-forward-det}. In addition, since FLUX includes a pair of Pack-Unpack operations around the diffusion transformer $f_\theta$, we employ a local continuity module (LCM) $\Lambda$ to mitigate grid artifacts caused by this Unpack operation.
    % there is a pair of Pack-Unpack operations around the diffusion transformer $f_\theta$ inherited from FLUX, a local continuity module (LCM) $\Lambda$ is employed to mitigate grid artifacts caused by this Unpack operation. 
    }
    \label{fig:1step}
\end{figure*}
\begin{figure}
    \centering
    \includegraphics[width=0.99\linewidth]{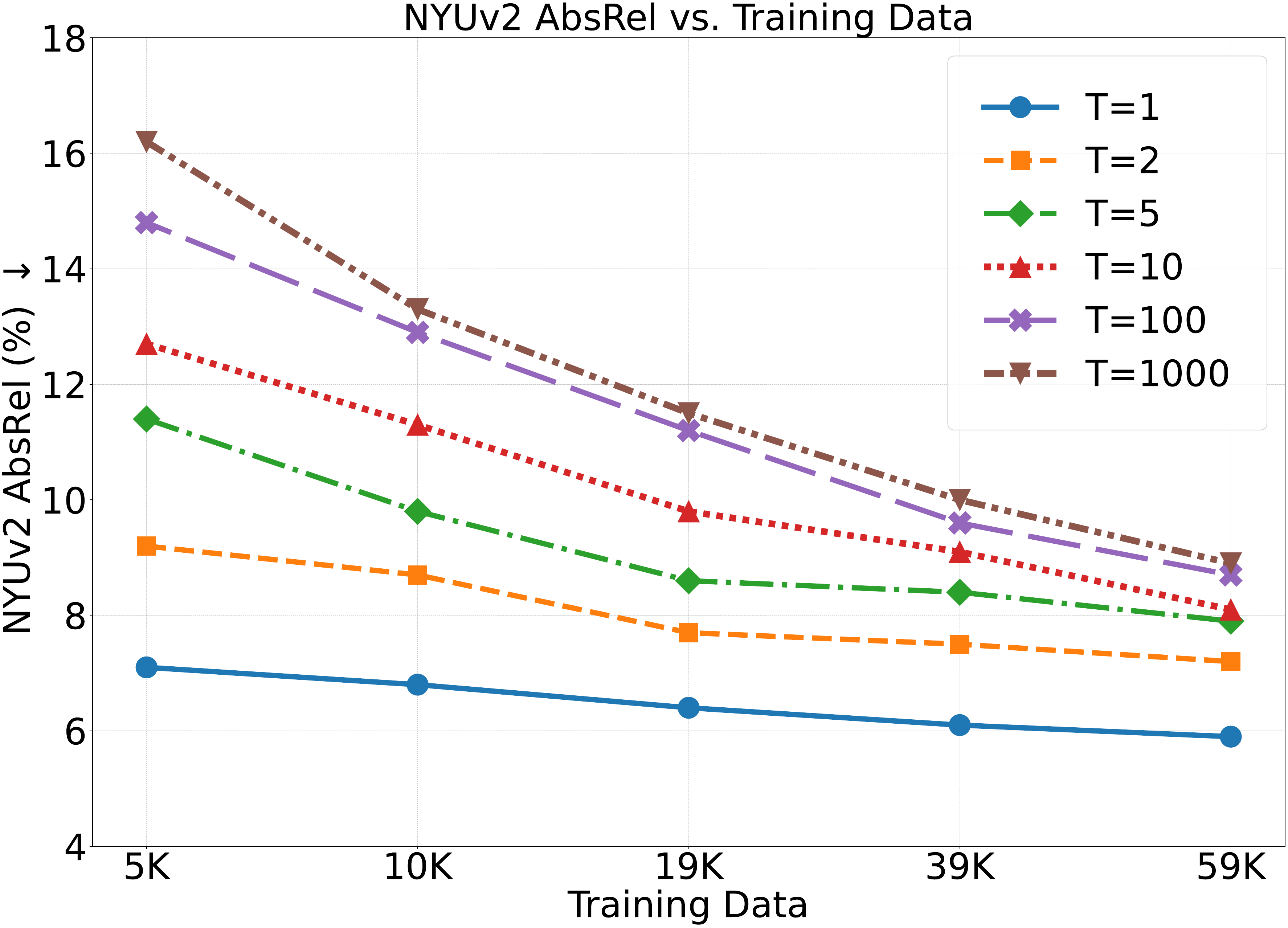}
    \caption{\textbf{Comparisons among various training time-steps and data scales} evaluated on NYUv2 in depth estimation. During inference, if the number of training time-steps $T>50$, the inference time-steps are fixed at $T_{\text{inf}}=50$; otherwise, $T_{\text{inf}}=T$.
    The results show that, when adapting the pre-trained rectified-flow model to dense prediction, reducing the number of training time-steps leads to improved performance. In particular, the single-step formulation ($T=1$) achieves the best performance across all data scales. }
    \label{fig:time}
\end{figure}
As illustrated in Fig.~\ref{fig:time}, we conduct experiments by gradually reducing the number of training time-steps $T$. This is achieved by modifying the value of $T$ in Eq. ~\ref{eq:time-set} to define new, smaller time-step sets for training. The results clearly show that the performance gradually improves as the number of
time-steps $T$ is reduced, culminating in the best result when reduced to only a single step. Under a stricter setting with more limited training data, the multi-step formulation is more sensitive to variations in training data scale compared to the single-step formulation. 
% , which is consistent with the observations in Lotus~\cite{he2024lotus}. 
The single-step formulation demonstrates greater stability and yields lower prediction errors. While it is plausible that, given unlimited high-quality data, both multi- and single-step formulations could reach comparable performance, such a setting is often costly and impractical for dense prediction tasks. 

Reducing the number of training time steps $T$ constrains the optimization space of rectified-flow formulation, thereby enabling more effective and efficient adaptation for geometric dense prediction. Motivated by this observation, we adopt the single-step formulation ($T=1$, \emph{i.e.}, $t=1$ in Eq.~\ref{eq:time-set}). This single-step formulation further enhances computational efficiency.

\subsubsection{Analysis-3, Parameterization Types}
\label{sssec:param}
Under the single-step formulation derived above, the model degenerates into a regression task, which is trained to predict the velocity given the input image with a fixed time-step $t=1$. The velocity $\mathbf{v} = \mathbf{z^x} - \mathbf{z^y}$ is the residual between the input image $\mathbf{z^x}$ and its annotation $\mathbf{z^y}$. We refer to this parameterization as \emph{Residual Prediction}.
During inference, the final prediction is obtained using the single-step Euler solver:
\begin{equation}
\label{eq:rf-inf}
    \mathbf{\hat{z}^y} = \mathbf{z^x} - f_\theta(\mathbf{z^x}, t),  
\end{equation}
where $f_\theta(\mathbf{z^x}, t)$ is the predicted residual. 

However, such residual prediction is problematic for dense prediction tasks for two reasons: \ding{172} Predicting $\mathbf{z^x}-\mathbf{z^y}$ requires the model to simultaneously learn image reconstruction and geometric estimation, which belong to substantially different distributions. This increases optimization difficulty and ultimately degrades accuracy; 
\ding{173} The predicted residual is dominated by high-frequency appearance signals of the input image, such as textures, illumination, and color. 
Although the term $\mathbf{z^x}$ in Eq.~\ref{eq:rf-inf} attempts to remove these appearance components during inference, however, imperfect prediction makes this removal unreliable, and appearance interference inevitably leaks into the final result. 

To overcome these limitations and better exploit pre-trained visual priors, we propose fine-tuning the model with \emph{Clean-Data Prediction}, \emph{i.e.}, directly predicting the clean annotation $\mathbf{z^y}$. 
The clean-data prediction offers a simpler and more direct training objective, alleviates optimization difficulty, and eliminates appearance interference, thereby yielding superior performance. 
\begin{figure}
    \centering
    \includegraphics[width=0.99\linewidth]{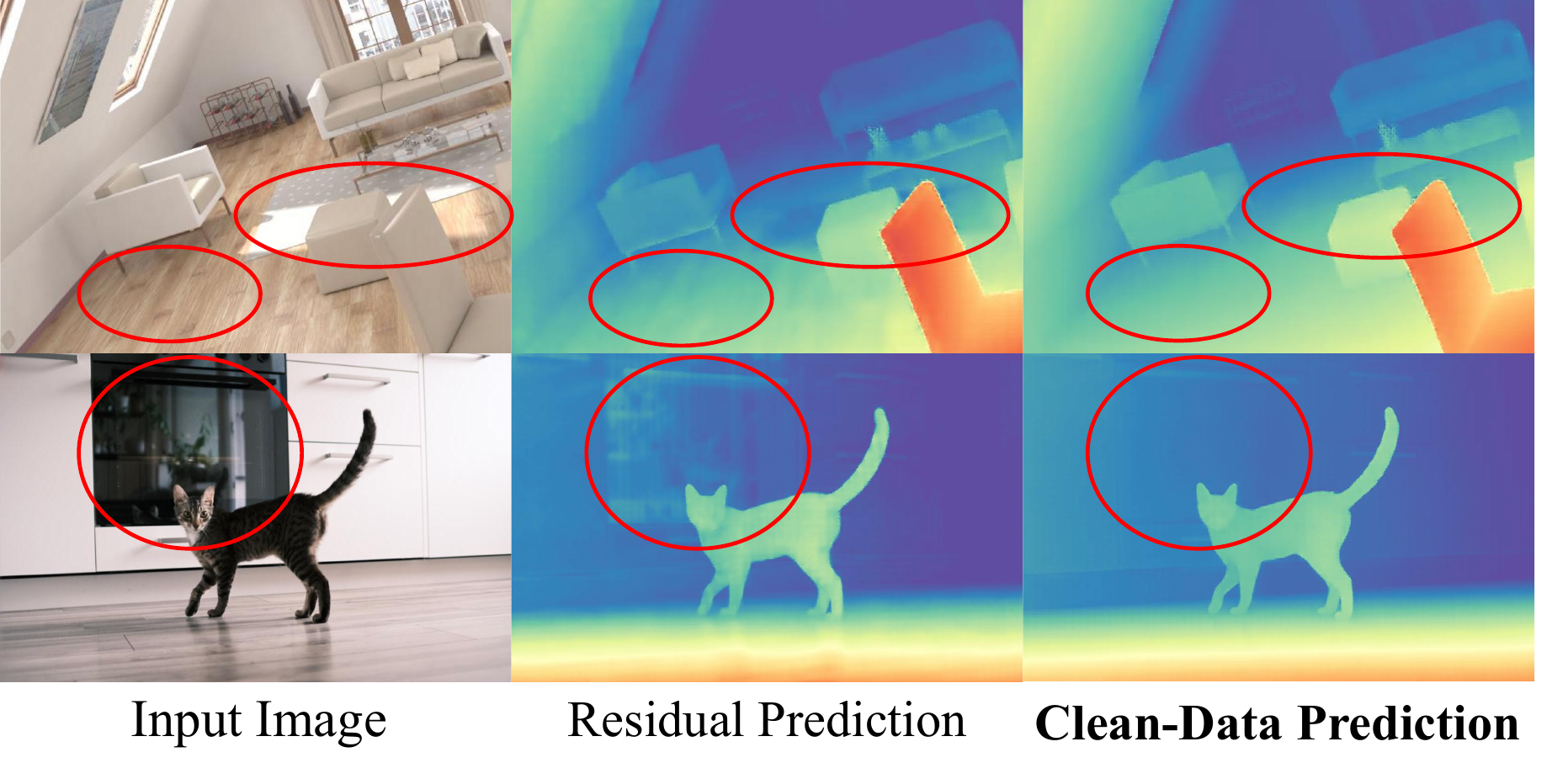}
    \caption{\textbf{Predictions under Different Model Parameterization Types.} \textcolor{red}{Red circles} highlight regions with obvious appearance artifacts when residual prediction is used. In contrast, clean-data prediction produces more accurate predictions without interference from image appearance. }
    \label{fig:flow-vs-x0}
\end{figure}

As shown in Fig.~\ref{fig:flow-vs-x0}, residual prediction produces predictions corrupted by image patterns (see red circles), whereas clean-data prediction yields accurate results without such interference. Consistently, Tab. \ref{tab:ablation} shows that clean-data prediction achieves significantly higher accuracy than the original residual prediction. 
Therefore, to mitigate appearance interference and improve prediction quality, we adopt clean-data prediction as the parameterization type. 

% Based on the above analysis, the core predictor employs the single-step formulation with clean-data prediction, as illustrated in Fig.~\ref{fig:1step}. The training objective becomes: 
% \begin{equation}
%     L_t = || \mathbf{z^y}-\Lambda(f_\theta(\mathbf{z_t}, t))||^2,
% \end{equation}
% where $t=1$ and the input latent $\mathbf{z_t}=\mathbf{z_1}=\mathbf{z^x}$ according to Eq.~\ref{eq: direct-rf-forward-det}. 

% \noindent \textbf{The Core Predictor Philosophy.} The systematic analyses in Sec. \ref{sssec:s_vs_d} to Sec. \ref{sssec:param} lead to a crucial insight: the key benefit of diffusion models for deterministic tasks lies not in their complex parameterization type or stochastic multi-step sampling process, but in the powerful world modeling capability embedded within their pre-trained weights. By building upon the robust deterministic formulation and dismantling the generative components—replacing them with the single-step formulation and the clean-data prediction—we transform the complex generative model into a highly efficient geometric regression. This strategy allows Lotus-2 to efficiently exploit the structural knowledge of the large-scale prior without inheriting its inherent instability and inefficiency. 

\subsubsection{Analysis-4, Local Continuity}
\label{sssec:lcm}
\begin{figure}[!t]
    \centering
    \includegraphics[width=0.99\linewidth]{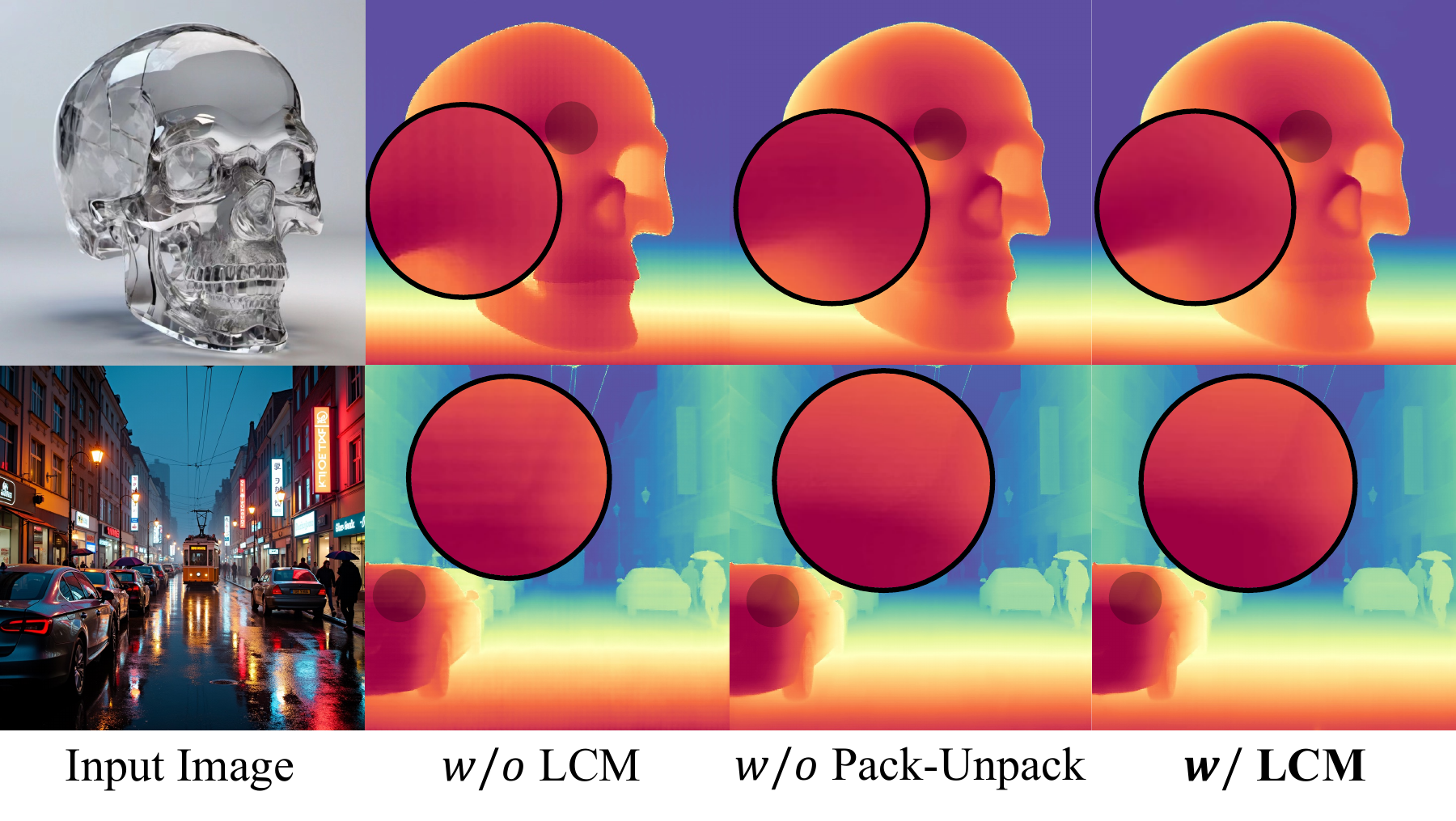}
    \caption{\textbf{The effects of different strategies for eliminating grid-like artifacts.} ``$w/o$ LCM'' refers to only single-step formulation with clean-data prediction, which produces noticeable grid-like artifacts due to the discontinuity introduced by Pack-Unpack.  Removing Pack-Unpack entirely alleviates this issue but compromises both accuracy and efficiency. In contrast, LCM effectively resolves the artifacts while improving accuracy and preserving model efficiency. (\textbf{Zoom in for clearer observation.}) }
    \label{fig:lcm}
\end{figure}

The FLUX architecture employs non-parametric \emph{Pack} and \emph{Unpack} operations to reduce computational overhead in the latent space for image generation (Sec.~\ref{sssec:pre-pack}). While efficient, the non-parametric nature of the Unpack operation, which rearranges feature channels back to spatial resolution after the diffusion transformer model, introduces spatial discontinuities at the boundaries of the $2\times 2$ latent patches. This localized discontinuity, which lacks constraints on local spatial coherence, is detrimental to geometric fidelity in the final output (Fig. ~\ref{fig:lcm}, ``$w/o$ LCM'').

To address this issue without compromising efficiency, we propose the lightweight \emph{Local Continuity Module} (LCM) after the Unpack operation of diffusion transformer backbone, as shown in Fig.~\ref{fig:1step}. LCM consists of two $3\times3$ convolutional layers with an intermediate GELU activation~\cite{hendrycks2016gaussian} to introduce nonlinearity, which is formally defined as:  
\begin{equation}
    \mathbf{\hat{z}^y} = \Lambda\big(f_\theta(\mathbf{z_t}, t)\big),
    \quad \Lambda(h) = \phi_2 \circledast \gamma(\phi_1 \circledast h),
\end{equation}
where $\Lambda(\cdot)$ denotes the LCM, $\circledast$ is the convolution operator, $\phi_1$ and $\phi_2$ are convolutional kernels, and $\gamma(\cdot)$ is the GELU activation.  

As shown in Fig .\ref{fig:lcm}, LCM effectively mitigates the local discontinuities introduced by Pack-Unpack, thereby eliminating grid artifacts. Furthermore, Tab.~\ref{tab:ablation} demonstrates that LCM not only improves visual quality but also enhances prediction accuracy. 

For comparison, we additionally evaluate a straightforward alternative: entirely removing the Pack-Unpack operations from FLUX architecture. While the removal of Pack-Unpack does eliminate grid artifacts (see the ``$w/o$ Pack-Unpack'' cases in Fig.~\ref{fig:lcm}), this approach suffers from two severe drawbacks: 
\ding{172} since the input–output dimensionality of the diffusion transformer changes, additional linear layers are required to align the dimensions, which shifts the feature space away from the pre-trained priors, degrading the prediction accuracy (see Tab.~\ref{tab:ablation}); 
\ding{173} the absence of Pack-Unpack drastically compromises model efficiency, leading to much slower inference speed.
Therefore, LCM offers an effective solution to the local discontinuity problem, while preserving the pre-trained priors and maintaining model efficiency. 

\subsubsection{Finalized Architecture and Objective}
The final core predictor is built upon the foundational Deterministic-DA and integrates all derived components: the single-step formulation, the clean-data prediction, and the local continuity module (LCM), as shown in Fig.~\ref{fig:1step}. This comprehensive design transforms the unstable and iterative generative flow into a highly efficient and structurally robust formulation, optimizing for deterministic geometric dense prediction. The overall training objective is defined as: 
\begin{equation}
     L_t = || \mathbf{z^y}-\Lambda(f_\theta(\mathbf{z_t}, t))||^2,
\end{equation}
where $t=1$ and the input latent $\mathbf{z_t}=\mathbf{z_1}=\mathbf{z^x}$. 

\subsection{Detail Sharpener: High-Fidelity Geometric Refinement}
\label{ssec:ds}

\begin{figure}[!ht]
    \centering
    \caption{\textbf{Comparisons in Detail Sharpness.} ``$w/o$ Sharpener'' denotes predictions directly obtained by the core predictor, which suffer from blurry and coarse details. The ``$w/$ Sharpener'' cases demonstrate that the detail sharpener noticeably enhances the sharpness of fine-grained structures, particularly along boundaries, while avoiding the  geometric hallucinations observed in Deterministic-DA, such as the misaligned chair backrest and stair railing. 
    % In this experiment, we set $T_{\text{inf}}'=T'=10$. 
    % (\textbf{Zoom in for clearer observation.} s) 
    }
    \includegraphics[width=0.99\linewidth]{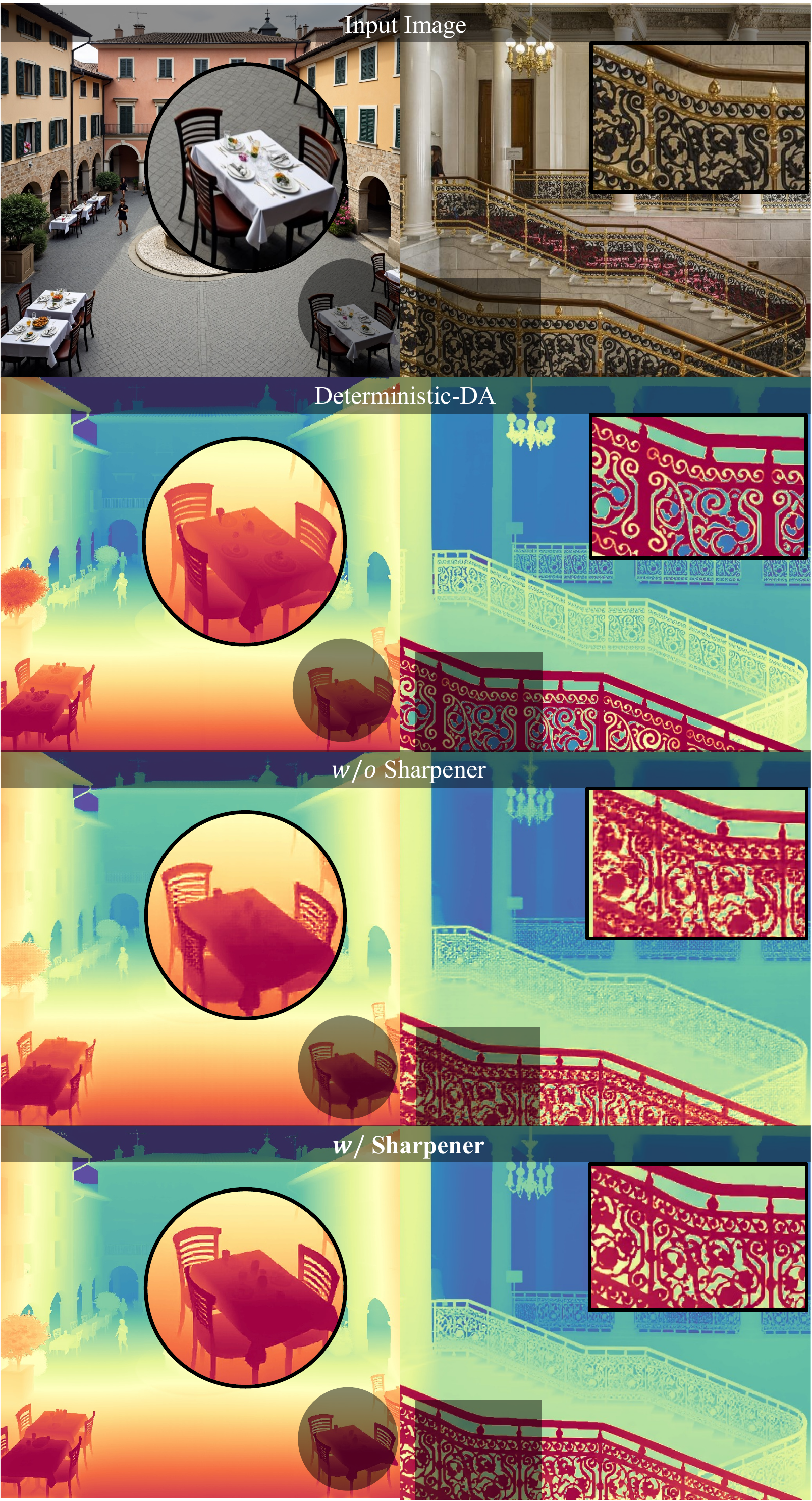}
    \label{fig:ds}
\end{figure}

The single-step core predictor excels at predicting accurate and globally coherent structure, but often produces predictions that are coarse and blurry in  high-frequency detail areas, lacking fine-grained fidelity (see ``$w/o$ Sharpener'' cases of Fig. ~\ref{fig:ds}). 
This limitation stems from the inherent difficulty of the single-step formulation in resolving high-frequency details. 
In contrast, multi-step flow (\emph{e.g.}, Deterministic-DA) retains the complexity to model high-frequency dynamics and can produce sharper details; however, due to its optimization difficulty and the accumulation of high errors across multiple steps, it is prone to geometric hallucination (see the ``Deterministic-DA'' cases of Fig.~\ref{fig:ds}), sacrificing structural correctness and overall accuracy.

\begin{figure*}[t]
    \centering
    \includegraphics[width=0.99\linewidth]{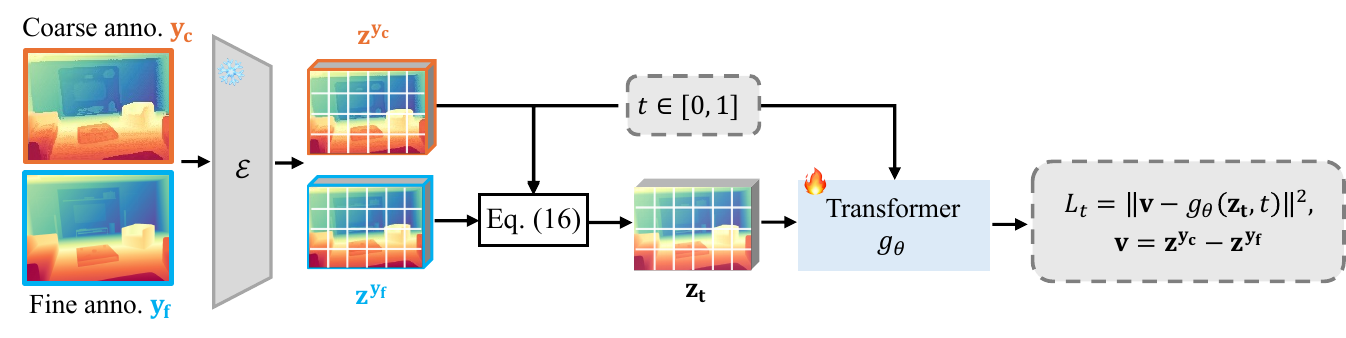}
    \caption{\textbf{The training pipeline of detail sharpener.} Starting from a structurally correct but coarse annotation predicted by the core predictor, the detail sharpener learns the transition from coarse to fine-grained annotation via a constrained multi-step rectified-flow within the manifold defined by the core predictor. 
    % We set the number of training time-steps $T'=10$ to balance the optimization efficiency and refinement quality, compromising the accuracy. 
    % The detail sharpener is fine-tuned from the pre-trained FLUX model to effectively leverage its visual priors for recovering fine details. 
    }
    \label{fig:ds-training}
\end{figure*}
\begin{figure*}
    \centering
    \includegraphics[width=0.99\linewidth]{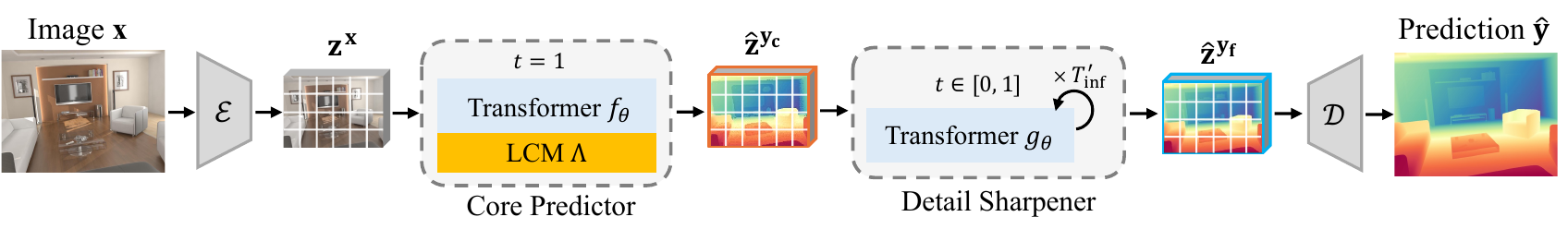}
    \caption{\textbf{The inference pipeline of Lotus-2.} It is a decoupled, two-stage deterministic pipeline that bridges the regression and geometric refinement. First, the core predictor produces stable and structurally consistent prediction via single-step regression. The detail sharpener then employs a constrained multi-step rectified-flow formulation for iterative refinement without any stochastic noise. The refinement uses $T_{\text{inf}}' \leq 10$ steps, adjustable based on the desired level of sharpness. This design ensures both structural consistency and fine-grained fidelity in minimal steps. }
    \label{fig:inf}
\end{figure*}
To simultaneously achieve accuracy and fine-grained fidelity, we introduce the \emph{Detail Sharpener}, a constrained multi-step rectified-flow model designed solely for geometric refinement within the manifold defined by the core predictor. 
Specifically, we first obtain a structurally correct but coarse prediction via the single-step core predictor, and then employ detail sharpener to progressively refine the high-frequency details. With this design, structural correctness is guaranteed by the core predictor, while the detail sharpener is solely responsible for enhancing sharpness. 

As illustrated in Fig.~\ref{fig:ds-training}, 
the detail sharpener is trained to learn a noise-free rectified-flow transformation from a coarse prediction $\mathbf{z^{y_c}}$ to its high-fidelity ground-truth $\mathbf{z^{y_f}}$. 
The flow is defined between the two known geometric states: 
\begin{equation}
    \mathbf{z_t} = t\mathbf{z^{y_c}}+(1-t)\mathbf{z^{y_f}}. 
\end{equation}
The model $g_\theta$ is fine-tuned from FLUX to predict the velocity $\mathbf{v} = \mathbf{z^{y_c}}-\mathbf{z^{y_f}}$. Thus, the training objective of detail sharpener is defined as: 
\begin{equation}
    L_t={||(\mathbf{z^{y_c}}-\mathbf{z^{y_f}})-g_\theta(\mathbf{z_t}, t)||}^2. 
    \label{eq:ds-training}
\end{equation} 
% Since the refinement primarily involves sharpness along boundaries, training on a limited dataset is sufficient to recover fine-grained fidelity. 
% Moreover, w
We set the number of training steps $T'=10$ to balance optimization and refinement. During inference, the number of inference steps $T_{\text{inf}}'$ can be flexibly chosen up to $T'$, depending on the desired level of sharpness. 

As shown in Fig.~\ref{fig:ds}, the detail sharpener noticeably enhances the sharpness while successfully avoiding the structural hallucinations observed in Deterministic-DA. Tab.~\ref{tab:ablation} further confirms that incorporating the detail sharpener does not compromise geometric accuracy established by the core predictor. 

\subsection{Inference}
\label{ssec:inference}
Lotus-2 executes a two-stage deterministic inference pipeline, as illustrated in Fig. ~\ref{fig:inf}. The core predictor is dedicated to ensuring structural correctness and efficiency, while the detail sharpener is solely responsible for high-fidelity refinement. 
Rooted in our philosophy of deterministic modeling, both the core predictor and the detail sharpener are noise-free, guaranteeing structural consistency and stability for deterministic geometric dense prediction.  
% Note that the detail sharpener is optional, requiring $T_{\text{inf}}' \leq 10$ steps based on the desired level of sharpness. 
The complete inference process proceeds as follows: 

\begin{enumerate}

\item The input image $\mathbf{x}$ is first encoded into the VAE latent space using the encoder $E$, yielding the image latent $\mathbf{z^x}$. 

\item The image latent $\mathbf{z^x}$ is passed through the core predictor to generate the accurate but coarse prediction $\mathbf{\hat{z}^{y_c}}$.  This step guarantees global structural correctness and is performed with maximum efficiency (1 step). 

 \item The coarse prediction $\mathbf{\hat{z}^{y_c}}$ is then fed into the detail sharpener to obtain the sharp and high-fidelity result $\mathbf{\hat{z}^{y_f}}$. 
 % , a constrained multi-step rectified-flow model that transforms the coarse latent into the sharp, high-fidelity result $\mathbf{\hat{z}^{y_f}}$. 
 This iterative refinement is achieved by the discrete Euler solver (Eq. ~\ref{eq:euler}). 
 Note that this refinement is optional based on the desired level of sharpness. 

\item The final refined latent $\mathbf{\hat{z}^{y_f}}$ is decoded back to the pixel space using the VAE decoder $D$ to produce the final geometric prediction $\mathbf{\hat{y}}$.

\end{enumerate} 

% This decoupled, deterministic approach enables Lotus-2 to achieve both the structural robustness required by geometric dense prediction and the high fidelity needed for fine-grained results, requiring only up to 11 total inference steps ($1 + 10$), which is significantly faster than prior stochastic generative methods. 

%% file: sec/4-exp_v2.tex
\section{Experiments}
In this section, we systematically validate the design principles of Lotus-2: leveraging pre-trained generative priors as a stable and deterministic flow for structurally correct and high-fidelity geometric dense prediction. We first detail the experimental setup, then present a quantitative comparison against state-of-the-art methods, followed by comprehensive ablation studies validating our methodological contributions. 
\subsection{Experimental Settings}
\subsubsection{Implementation Details}
We implement the proposed Lotus-2, which includes both the core predictor and the detail sharpener, by fine-tuning the pre-trained FLUX model~\cite{bfl2024flux} without utilizing the text conditioning. 
Our design adapts the rectified-flow formulation by setting the core predictor to a single-step formulation ($T=1$, $t=1$) with clean-data prediction and the detail sharpener to a constrained multi-step rectified-flow formulation ($T'=10$, with time-steps defined by Eq. \ref{eq:time-set}) within the manifold defined by the core predictor. 
For optimization, we use the Adam optimizer with a learning rate of $1\times10^{-4}$. 
All models are trained on 8 NVIDIA H100 GPUs (80G) with a total batch size of 64. 
To adapt the large-scale pre-trained architecture, we employ the parameter-efficient method LoRA~\cite{hu2022lora}, using a rank of 128 for depth estimation and 256 for normal estimation. For depth estimation, we operate in the disparity space, \textit{i.e.}, $d = 1/d'$, where $d'$ is the true depth. 
During inference, the core predictor directly predicts the coarse but structurally correct prediction in a single inference step, while the detail sharpener utilizes the Euler sampler with $T_{\text{inf}}'=T'=10$ steps for refinement. 
\subsubsection{Training Datasets}
A core demonstration of this work is the ability to achieve SoTA performance using extremely limited supervised data. Both depth and normal estimation tasks are trained solely on a small collection of synthetic data, totaling approximately \textbf{59K samples}—a fraction of the millions used by large-scale discriminative models. 
\begin{itemize}
\item \emph{Hypersim}~\cite{roberts2021hypersim}: A photorealistic synthetic dataset of 461 indoor scenes. We utilize the official training split, retaining approximately 39K samples after filtering. All samples are resized to $576 \times 768$.
\item \emph{Virtual KITTI} (VKITTI)~\cite{cabon2020virtual}: A synthetic street-scene dataset covering five urban scenes. We use four scenes, comprising about 20K samples, cropped to $352 \times 1216$. \end{itemize} 
To train the detail sharpener, we implement the methodology described in Sec.~\ref{ssec:ds}: we first generate coarse predictions ($\mathbf{y_c}$) on Hypersim and VKITTI using the trained core predictor, and then train the detail sharpener on the flow defined between these coarse predictions and the ground truth ($\mathbf{y_f}$). 
\subsubsection{Evaluation Datasets}
\label{ssec:eval-dataset}
We evaluate the generalization capability of Lotus-2 across five real-world datasets for depth estimation and four for surface normal prediction, none of which were seen during training. 
\begin{itemize} \item \emph{Affine-Invariant Depth Estimation}: We evaluate across diverse indoor (NYUv2~\cite{silberman2012indoor}, ScanNet~\cite{dai2017scannet}), outdoor (KITTI~\cite{geiger2013vision}), and high-resolution mixed scenes (ETH3D~\cite{schops2017multi}, DIODE~\cite{vasiljevic2019diode}). \item \emph{Surface Normal Prediction}: We use NYUv2, ScanNet, and iBims-1~\cite{koch2018evaluation} for real indoor evaluation, and Sintel~\cite{butler2012naturalistic} for highly dynamic synthetic outdoor scenes. \end{itemize}
\subsubsection{Metrics}
\label{sec:metrics}
We employ widely accepted metrics for both affine-invariant depth estimation and surface normal prediction. 
\begin{itemize}
    \item \emph{Affine-Invariant Depth Estimation}: 
    Following standard protocols~\cite{ranftl2020towards,ke2024repurposing}, we firstly align predictions to ground truth via least-squares fitting before evaluation. We report two primary metrics: the \emph{absolute mean relative error} (AbsRel), defined as $\frac{1}{M}\sum_{i=1}^M |a_i-d_i|/d_i$ (lower is better); and the $\delta 1$ value, which is the proportion of pixels satisfying $\text{Max}(a_i/d_i, d_i/a_i)<1.25$ (higher is better). 
    \item \emph{Surface Normal Prediction}: Following~\cite{bae2024dsine,ye2024stablenormal}, we measure the \emph{mean angular error} (lower is better) and the percentage of pixels with an angular error below $11.25^{\circ}$ (higher is better). 
\end{itemize}
For overall comparison, we report the \emph{Avg. Rank}, which is the average ranking of each method across all datasets and metrics. A lower Avg. Rank signifies superior overall performance. 
\setlength{\tabcolsep}{2.pt}
\begin{table*}[!t]
% \vspace{-5mm}
\scriptsize
\caption{\textbf{Quantitative comparison on zero-shot affine-invariant depth estimation} between Lotus-2 and SoTA methods. 
% The upper section lists discriminative methods, the lower lists generative ones.
The \colorbox{best}{best} and \colorbox{best2}{second best} performances are highlighted. 
$^\S$indicates results re-evaluated by ourselves using the evaluation protocol of Marigold~\cite{ke2024repurposing}. $^\star$denotes the method relies on pre-trained text-to-image generative models. 
Our Lotus-2 achieves the best overall performance among all methods.
% Note that we only use $0.66\%$ of the training data compared to MoGe series and $0.09\%$ compared to Depth Anything series. 
}
% \vspace{-3mm}
\label{tab:depth}
\scriptsize

\resizebox{\textwidth}{!}{
\begin{tabular}{l|c|cc|cc|cc|cc|cc|c}
\toprule

\multirow{2}{*}{Method} 
& Training
& \multicolumn{2}{c|}{NYUv2 (Indoor)} 
& \multicolumn{2}{c|}{KITTI (Outdoor)} 
& \multicolumn{2}{c|}{ETH3D (Various)} 
& \multicolumn{2}{c|}{ScanNet (Indoor)} 
& \multicolumn{2}{c|}{DIODE (Various)} 
& Avg.\\

 & Data$\downarrow$
 & AbsRel$\downarrow$ & $\delta$1$\uparrow$  
 & AbsRel$\downarrow$ & $\delta$1$\uparrow$  
 & AbsRel$\downarrow$ & $\delta$1$\uparrow$  
 & AbsRel$\downarrow$ & $\delta$1$\uparrow$ 
 & AbsRel$\downarrow$ & $\delta$1$\uparrow$
 & Rank       \\
\midrule

DiverseDepth 
& 320K
& 11.7 & 87.5 
& 19.0 & 70.4  
& 22.8 & 69.4 
& 10.9 & 88.2 
& 37.6 & 63.1
& 19.5            \\

MiDaS 
&2M
& 11.1 & 88.5 
& 23.6 & 63.0 
& 18.4 & 75.2 
& 12.1 & 84.6 
& 33.2 & 71.5
& 18.7               \\

LeRes 
&354K
& 9.0 & 91.6 
& 14.9 & 78.4 
& 17.1 & 77.7 
& 9.1 & 91.7 
& 27.1 & 76.6
& 15.7              \\

Omnidata
&12.2M
& 7.4 & 94.5 
& 14.9 & 83.5 
& 16.6 & 77.8 
& 7.5 & 93.6 
& 33.9 & 74.2
& 15.4 \\ 

DPT
&1.4M
& 9.8 & 90.3 
& 10.0 & 90.1 
& 7.8 & 94.6 
& 8.2 & 93.4 
& 18.2 & 75.8 
& 12.5            \\ 

GeoWizard$^{\star^\S}$
& 280K
& 5.6 & 96.3 
& 14.4 & 82.0 
& 6.6 & 95.8
& 6.4 & 95.0 
& 33.5 & 72.3
& 12.4          \\

HDN
&300K
& 6.9 & 94.8 
& 11.5 & 86.7 
& 12.1 & 83.3 
& 8.0 & 93.9 
& 24.6 & \cellcolor{best2}78.0 
& 12.2             \\ 

GenPercept$^{\star^\S}$       
& 74K
& 5.6 & 96.0 
& 13.0 & 84.2 
& 7.0 & 95.6 
& 6.2 & 96.1 
& 35.7 & 75.6
& 11.5           \\ 

Marigold$_\text{(LCM)}$$^{\star\S}$
& 74K
& 6.1 & 95.8 
& 9.8 & 91.8 
& 6.8 & 95.6
& 6.9 & 94.6 
& 30.7 & 77.5
& 10.5           \\

MoGe-2$^{\S}$
& 8.9M
& \cellcolor{best}3.6 & \cellcolor{best2}98
& 11.8 & 89.2
& 16.6 & 81.5
& \cellcolor{best}3.5 & \cellcolor{best2}98.2
& 39.3 & 70.0
& 10.4          \\

Marigold$^{\star}$
& 74K
& 5.5 & 96.4
& 9.9 & 91.6  
& 6.5 & 95.9 
& 6.4 & 95.2 
& 30.8 & 77.3
& 9.2            \\

DICEPTION$^{\star}$
& 500K
& 7.2 & 93.9
& 7.5 & 94.5
& \cellcolor{best2}5.3 & 96.7
& 7.5 & 93.8
& 24.3 & 74.1
& 9.2            \\

DepthAnything V2
& 62.6M
& 4.5 & 97.9
& \cellcolor{best2}7.4 & \cellcolor{best2}94.6
& 13.1 & 86.5
& 4.2 & 97.8
& 26.5 & 73.4
& 7.3 \\

Diffusion-E2E-FT$^{\star}$
& 74K
& 5.4 & 96.5
& 9.6 & 92.1
& 6.4 & 95.9
& 5.8 & 96.5
& 30.3 & 77.6
& 7.1 \\

Lotus-G$^{\star}$
& \cellcolor{best}59K 
& 5.4 & 96.8
& 8.5 & 92.2
& 5.9 & \cellcolor{best2}97.0
& 5.9 & 95.7
& 22.9 & 72.9
& 7.1 \\

DepthFM-ID$^{\star}$
& 81.4K
& 5.5 & 96.3
& 8.9 & 91.3
& 5.8 & 96.2
& 6.3 & 95.4
& \cellcolor{best}21.2 & \cellcolor{best}80.0
& 6.9         \\

MoGe$^{\S}$
& 9M
& \cellcolor{best}3.6 & 97.9
& 7.3 & 95.2
& 8.4 & 93.0
& \cellcolor{best}3.5 & \cellcolor{best}98.4
& 36.3 & 71.2
& 6.9          \\

DepthAnything
& 62.6M
& 4.3 & \cellcolor{best}98.1
& 7.6 & \cellcolor{best}94.7
& 12.7 & 88.2
& 4.3 & 98.1
& 26.0 & 75.9
& 6.2  \\

\cellcolor{best2}Lotus-D$^{\star}$ 
& \cellcolor{best}59K
& 5.1 & 97.2
& 8.1 & 93.1
& 6.1 & 97.0
& 5.5 & 96.5
& 22.8 & 73.8 
& \cellcolor{best2}6.0  \\

\cellcolor{best}\textbf{Lotus-2}$^{\star}$ 
& \cellcolor{best}\textbf{59K}
& \cellcolor{best2}\textbf{4.1} & \textbf{97.6}
& \cellcolor{best}\textbf{6.7} & \textbf{94.5}
& \cellcolor{best}\textbf{4.6} & \cellcolor{best}\textbf{98.1}
& \cellcolor{best2}\textbf{4.2} & \textbf{97.6}
& \cellcolor{best2}\textbf{22.1} & \textbf{75.2}
& \cellcolor{best}\textbf{3.6}         \\

\bottomrule
\end{tabular}
}
\end{table*}

\setlength{\tabcolsep}{5pt}
\begin{table*}[!h]
% \vspace{-1mm}
\scriptsize
% \vspace{-7mm}
\caption{\textbf{Quantitative comparison on zero-shot surface normal estimation} between Lotus-2 and SoTA methods. 
$^\ddag$refers the Marigold normal model as detailed in this \href{https://huggingface.co/prs-eth/marigold-normals-lcm-v0-1}{link}. 
$^\S$indicates results re-evaluated by us using the evaluation protocol of DSINE~\cite{bae2024dsine}. 
Our Lotus-2 demonstrates highly competitive quantitative performance, crucially delivering the robust and fine-grained qualitative results as highlighted in Fig.~\ref{fig:teaser}. 
}
% \vspace{-3mm}
\label{tab:normal}
\scriptsize
\centering
\resizebox{\textwidth}{!}{
\begin{tabular}{l|c|cc|cc|cc|cc|cc|c}
\toprule

\multirow{2}{*}{Method} 
& Training
& \multicolumn{2}{c|}{NYUv2 (Indoor)} 
& \multicolumn{2}{c|}{ScanNet (Indoor)} 
& \multicolumn{2}{c|}{iBims-1 (Indoor)} 
& \multicolumn{2}{c|}{Sintel (Outdoor)}
& \textcolor{black}{Avg.} \\
& Data$\downarrow$
& mean$\downarrow$ & $11.25^\circ$$\uparrow$
& mean$\downarrow$ & $11.25^\circ$$\uparrow$
& mean$\downarrow$ & $11.25^\circ$$\uparrow$
& mean$\downarrow$ & $11.25^\circ$$\uparrow$
& Rank            \\
\midrule

OASIS
& 110K
& 29.2 & 23.8 
& 32.8 & 15.4 
& 32.6 & 23.5 
& 43.1 & 7.0 
& 13.5            \\

Omnidata
& 12.2M
& 23.1 & 45.8 
& 22.9 & 47.4 
& 19.0 & 62.1 
& 41.5 & 11.4 
& 11.9           \\

GeoWizard$^{\star\S}$
& 280K
& 18.9 & 50.7 
& 17.4 & 53.8 
& 19.3 & 63.0 
& 40.3 & 12.3 
& 10.4               \\

StableNormal$^{\star\S}$
& 250K
& 18.6 & 53.5 
& 17.1 & 57.4 
& 18.2 & 65.0 
& 36.7 & 14.1 
& 8.4  \\

GenPercept$^{\star\S}$
& 74K
& 18.2 & 56.3 
& 17.7 & 58.3 
& 18.2 & 64.0 
& 37.6 & 16.2 
& 8.3            \\

EESNU
& 2.5M
& 16.2 & 58.6 
& -    & -     
& 20.0 & 58.5  
& 42.1 & 11.5  
& 7.3                 \\

Omnidata V2
& 12.2M
& 17.2 & 55.5 
& 16.2 & 60.2 
& 18.2 & 63.9 
& 40.5 & 14.7 
& 8.1                \\

Marigold$^{\star\ddag}$
& 74K
& 20.9 & 50.5 
& 21.3 & 45.6 
& 18.5 & 64.7 
& -  & -  
& 8.1           \\

Lotus-G$^\star$ 
& \cellcolor{best}59K
& 16.5 & 59.4 
& 15.1 & 63.9 
& 17.2 & 66.2 
& 33.6 & 21.0 
& 5.4  \\

DSINE
& 160K
& 16.4 & 59.6 
& 16.2 & 61.0 
& 17.1 & 67.4 
& 34.9 & 21.5 
& 4.9              \\

Diffusion-E2E-FT$^{\star\S}$
& 74K
& 16.5 & \cellcolor{best2}60.4 
& 14.7 & 66.1 
& 16.1 & \cellcolor{best2}69.7 
& 33.5 & 22.3  
& 3.4   \\

Lotus-D$^\star$ 
& \cellcolor{best}59K
& \cellcolor{best2}16.2 & 59.8 
& 14.7 & 64.0 
& 17.1 & 66.4
& 32.3 & 22.4 
& 3.4  \\

\cellcolor{best2}\textbf{Lotus-2}$^\star$ 
& \cellcolor{best}\textbf{59K}
& \textbf{16.9} & \textbf{59.0}
& \cellcolor{best2}\textbf{14.2} & \cellcolor{best2}\textbf{66.8}
& \cellcolor{best2}\textbf{15.4} & \cellcolor{best}\textbf{70.4}   
& \cellcolor{best2}\textbf{30.3} & \cellcolor{best}\textbf{27.6}
& \cellcolor{best2}\textbf{2.9}         \\

\cellcolor{best}MoGe-2$^\S$ 
& 8.9M
& \cellcolor{best}14.7 & \cellcolor{best}62.3
& \cellcolor{best}12.8 & \cellcolor{best}68.4
& \cellcolor{best}14.7 & \cellcolor{best}70.4
& \cellcolor{best}29.3 & \cellcolor{best2}24.8
& \cellcolor{best}1.1        \\

\bottomrule
\end{tabular}
}
\end{table*}

\subsection{Comparison with State-of-the-Art}
We benchmark Lotus-2 against recent state-of-the-art methods in both affine-invariant monocular depth estimation and surface normal prediction, including both large-scale discriminative models (\emph{e.g.}, DepthAnything~\cite{yang2024depth, yang2024depth2}, MoGe~\cite{wang2025moge, wang2025moge2}) and generative prior adaptation methods (\emph{e.g.}, Marigold~\cite{ke2024repurposing}, GeoWizard~\cite{fu2024geowizard}). 

\subsubsection{Affine-Invariant Depth Estimation}
As presented in Tab.~\ref{tab:depth},  Lotus-2 establishes a new state-of-the-art in affine-invariant monocular depth estimation across the five real-world datasets. Notably, Lotus-2 achieves the best Avg. Rank despite being trained on only 59K samples. This result decisively validates the power of leveraging large-scale generative models as deterministic world priors, allowing Lotus-2 to surpass massive data-trained discriminative methods. 

\subsubsection{Surface Normal Prediction}
For surface normal prediction, Lotus-2 demonstrates highly competitive performance (Tab.~\ref{tab:normal}), showcasing the effectiveness of our deterministic adaptation in capturing complex geometry. Crucially, as highlighted in Fig.~\ref{fig:teaser}, our deterministic adaptation of world priors ensures robust and structurally correct geometric prediction, enabling strong generalization even in challenging or rare scenes. This robust foundation, coupled with our noise-free multi-step refinement (detail sharpener), proves highly effective at capturing the high-frequency surface detail required for local geometry, significantly outperforming other SoTA approaches. 
% \hj{\subsection{Qualitative Comparisons}}
\setlength{\tabcolsep}{5.pt}
\begin{table*}[!t]
% \vspace{-3mm}
\scriptsize
% \vspace{-15mm}
\caption{\textbf{Ablation studies} of the proposed Lotus-2. The second portion of the table contains the key components of the \emph{core predictor}, sequentially demonstrating the performance gains conferred by each design. The final row validates the  \emph{detail sharpener}. The shaded row \textcolor{gray}{\colorbox{gray!20}{($w/o$ Pack-Unpack)}} is included as an auxiliary ablation to validate the effect of the local continuity module (LCM). The results below are evaluated in monocular depth estimation across four datasets. }
% \vspace{-3mm}
\label{tab:ablation}
\scriptsize
\resizebox{\textwidth}{!}{
\begin{tabular}{l|cc|cc|cc|cc}
\toprule
\multirow{2}{*}{Method} 
% & Training
& \multicolumn{2}{c|}{NYUv2 (Indoor)} 
& \multicolumn{2}{c|}{KITTI (Outdoor)} 
& \multicolumn{2}{c|}{ETH3D (Various)} 
& \multicolumn{2}{c}{ScanNet (Indoor)} \\
% & Data
& AbsRel$\downarrow$ & $\delta$1$\uparrow$
& AbsRel$\downarrow$ & $\delta$1$\uparrow$
& AbsRel$\downarrow$ & $\delta$1$\uparrow$ 
& AbsRel$\downarrow$ & $\delta$1$\uparrow$ 
\\
\midrule
Stochastic-DA   
% & 39K          
& 8.261 & 93.468 
& 13.196 & 78.204 
& 17.384 & 77.842 
& 9.373 & 91.569  \\
\midrule
Deterministic-DA
% & 59K          
& 7.812 & 94.262 
& 10.212 & 89.900
& 10.766 & 94.762
& 8.488 & 92.897  \\

$+$ Single-Step Formulation  
% & 59K
& 5.910 & 96.939
& 8.833 & 92.088
& 5.858 & 96.952
& 7.121 & 96.331  \\

$+$ Clean-Data Prediction  
% & 59K
& 4.384  & 97.627
& 6.843 & 94.325 
& 4.980 & 97.552 
& 4.446 & 97.529  \\

$+$ Local Continuity Module
% & 59K       
& 4.128 & 97.608
& 6.576  & 94.682
& 4.625 & 98.004 
& 4.174 & 97.575    \\

\rowcolor{gray!20} \color{gray}
\quad($w/o$ Pack-Unpack) 
& \color{gray}4.817 & \color{gray}97.383
& \color{gray}6.966 & \color{gray}94.203
& \color{gray}5.728 & \color{gray}97.252
& \color{gray}4.723 & \color{gray}97.168 \\ 
\midrule
$+$ Detail Sharpener
% & 59K       
& 4.122 & 97.623
& 6.725  & 94.492
& 4.643 & 98.101
& 4.188 & 97.597    \\

\bottomrule
\end{tabular}
}
\end{table*}
\subsection{Ablation Studies}
\subsubsection{Ablation on the Core Predictor}

The core predictor is the structural foundation of Lotus-2. We systematically validate its design in Tab.~\ref{tab:ablation} by incrementally incorporating the core contributions, showing consistent performance superiority across all four evaluation datasets. 

We begin by validating the necessity of the deterministic formulation. Moving from the stochastic generative formulation (Stochastic-DA) to the noise-free deterministic formulation (Deterministic-DA) yields an immediate improvement in accuracy. This validates our core hypothesis that deterministic geometric prediction requires a stable flow (Sec.~\ref{sssec:s_vs_d}). 
Next, adopting the single-step formulation ($T=1$) also provides a significant performance increase, confirming the single-step mechanism is the optimal strategy for efficiently leveraging pre-trained world priors under limited data (Sec.~\ref{sssec:num-timestep}). 
Following this, switching to clean-data prediction from residual prediction consistently achieves higher structural accuracy. This confirms that its value lies in both eliminating high-frequency appearance interference (Fig.~\ref{fig:flow-vs-x0}) and providing a more direct and effective optimization target (Sec.~\ref{sssec:param}). 
Finally, we validate the local continuity module (LCM). This lightweight module successfully eliminates grid artifacts (Fig.~\ref{fig:lcm}) and provides the final accuracy boost. This contrasts with the ``$w/o$ Pack-Unpack'' alternative, which compromises efficiency and degrades performance due to feature space misalignment (Sec.~\ref{sssec:lcm}).

\subsubsection{Ablation on the Detail Sharpener} 
% \begin{figure}
%     \centering
%     \includegraphics[width=0.99\linewidth]{imgs/plot_ds_t.pdf}
%     \caption{\textbf{The effect of different training time-steps $T'$ for detail sharpener. } The quantitative results are measured on NYUv2 dataset for monocular depth estimation. In all experiments,  the inference steps $T_{\text{inf}}'$ are determined by $T'$ using the rule: $T_{\text{inf}}'=50$ if $T'>50$, otherwise $T_{\text{inf}}'=T'$. The figure shows that $T'=10$ serves as the optimal inflection point, balancing optimization efficiency against sufficient capacity to achieve fine-grained fidelity. } 
%     \label{fig:ds_t}
% \end{figure}
The detail sharpener is responsible for high-fidelity refinement via a constrained multi-step flow. This ablation validates the contribution of the detail sharpener to high-fidelity geometry through qualitative results, quantitative metrics, and spectral analysis.

% \noindent\textbf{Qualitative and Quantitative Ablation.}: 
As qualitatively demonstrated in Fig.~\ref{fig:ds}, the detail sharpener achieves noticeable refinement in high-frequency areas. Quantitatively, the final line item in Tab.~\ref{tab:ablation} shows that the multi-step flow of the detail sharpener maintains the near-optimal accuracy achieved by the core predictor. This preservation of accuracy confirms that the detail sharpener successfully operates on a decoupled objective—enhancing local fidelity—without compromising the structural accuracy established by the core predictor, thus validating the success of our two-stage design.

\begin{figure}
    \centering
    \includegraphics[width=1.0\linewidth]{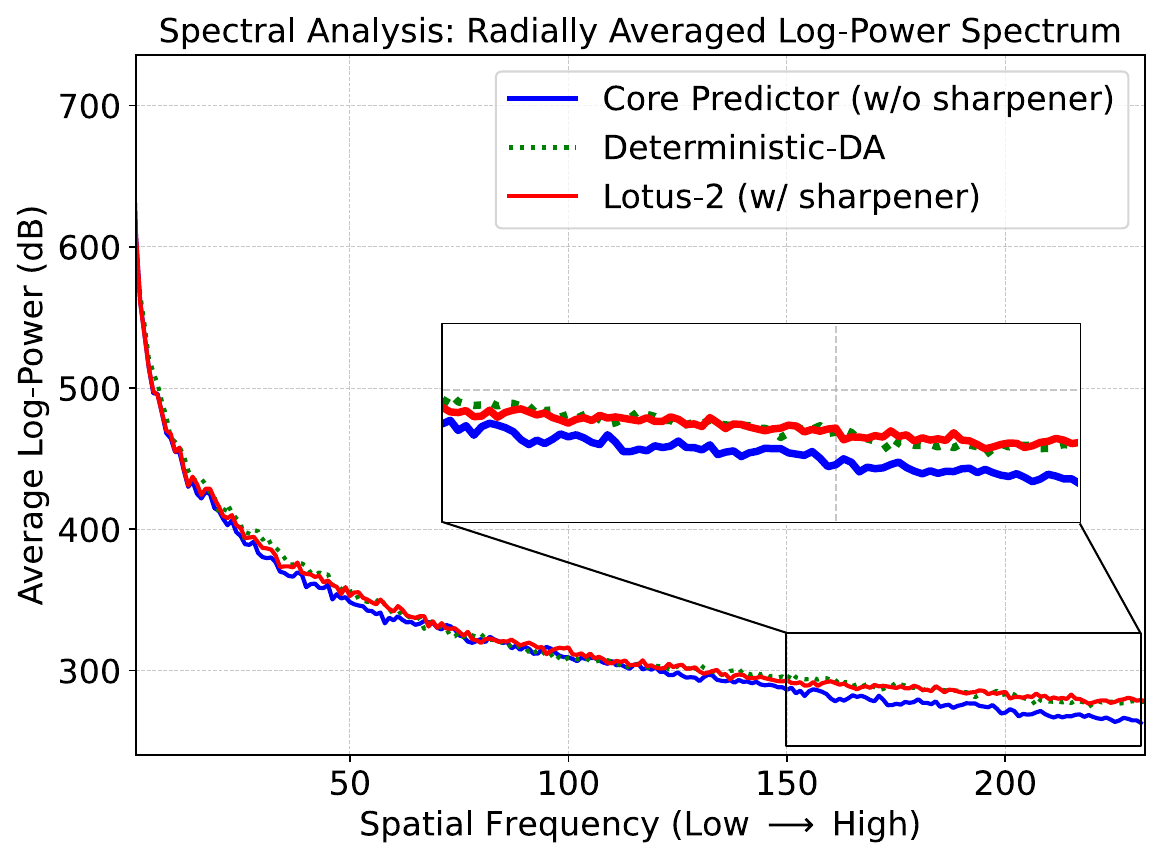}
    \caption{\textbf{Spectral analysis of high-fidelity refinement.} This plot compares the average log-power (y-axis) across spatial frequencies (x-axis) on NYUv2 dataset to validate the contribution of detail sharpener. The decay of the core predictor ($w/o$ sharpener) curve confirms its coarse nature, while the Lotus-2 ($w/$ sharpener) curve shows recovery of high-frequency power. }
    \label{fig:spectral}
\end{figure}
% \noindent\textbf{Spectral Analysis on fine-grained fidelity}: 
To rigorously quantify the contribution of the detail sharpener to fine-grained fidelity, specifically its effect on high-frequency detail areas, we conduct a spectral analysis using the 1D radially averaged power spectrum as illustrated in Fig.~\ref{fig:spectral}. The results show that the prediction from the core predictor exhibits a clear decay in power at high frequencies, confirming its output is structurally correct but coarse. In contrast, both Deterministic-DA and our Lotus-2 retain significantly more high-frequency power, indicating successful detail refinement. This provides quantitative, signal-level evidence that the detail sharpener is essential for high-fidelity geometric prediction.

% \noindent\textbf{Optimal Training Time-Steps ($T'$)}: To find the optimal strategy for modeling high-frequency dynamics, we perform an ablation study on the number of training time-steps $T'$. The selection criterion is maintaining high accuracy—as excessive steps hinder optimization as in Sec.~\ref{sssec:num-timestep}—while providing enough steps for effective refinement. As shown in Fig.~\ref{fig:ds_t}, accuracy sharply improves as $T'$ is reduced from $T'=1000$ to $T'=10$. Below $T'=10$ (\emph{i.e.}, $T'=5$ and $T'=2$), the quantitative gain becomes marginal. We select $T'=10$ as the optimal trade-off, which maximizes optimization efficiency while providing sufficient capacity (10 inference steps) for the detail sharpener to achieve fine-grained fidelity.

%% file: sec/5-conclusion_v2.tex
\section{Conclusion}
In this work, we addressed the fundamental challenge of geometric dense prediction—the task's ill-posed nature—by proposing a critical shift in how large-scale generative models are leveraged. We established the principle that for deterministic geometric inference, the power of diffusion backbones lies not in their stochastic sampling process but in their implicitly embedded deterministic world priors. Directly reusing the original stochastic generative flow proves suboptimal, leading to structural variance and unacceptable inconsistency in geometric outputs. 

To fully exploit these priors in a disciplined and stable manner, we introduced Lotus-2, a novel two-stage deterministic framework that decouples the inference process into two specialized, noise-free rectified-flow mappings. 

The first stage, the \emph{core predictor}, is implemented for maximum structural accuracy and efficiency. Through systematic ablation, we validated the necessity of our derived design choices: the deterministic shift, the highly efficient single-step formulation ($T=1$), and the clean-data prediction objective, which together transform the complex generative flow into a robust geometric regressor. The lightweight local continuity module (LCM) further ensures fidelity by suppressing architectural artifacts without compromising efficiency. 

The second stage, the \emph{detail sharpener}, solves the final limitation of single-step regression—coarse high-frequency details. It performs a constrained multi-step refinement within the geometry manifold established by the core predictor. This process is inherently noise-free and is optimized to selectively enhance high-fidelity geometry without compromising the established global structural correctness, successfully validating the benefits of our decoupled design. 

The experimental results decisively confirm our core hypothesis. By training on only 59K synthetic samples—less than $1\%$ of existing large-scale datasets—Lotus-2 achieved new state-of-the-art performance in monocular depth estimation and demonstrated highly competitive results in surface normal prediction. This unprecedented data efficiency, combined with high inference stability and fine-grained fidelity, validates the efficacy of our deterministic adaptation protocol. 

Ultimately, this work demonstrates that the vast knowledge accumulated by generative diffusion models can be repurposed to enable efficient, accurate, and physically consistent geometric reasoning, setting a new paradigm for structured prediction tasks beyond traditional discriminative and generative methods. This finding opens promising avenues for future research into extracting and utilizing structured knowledge from foundational generative models. 

%% file: sec/acknowledgments.tex
% \section*{Acknowledgments}
% This should be a simple paragraph before the References to thank those individuals and institutions who have supported your work on this article.

%% file: sec/appendix.tex
% {\appendix[Proof of the Zonklar Equations]
% Use $\backslash${\tt{appendix}} if you have a single appendix:
% Do not use $\backslash${\tt{section}} anymore after $\backslash${\tt{appendix}}, only $\backslash${\tt{section*}}.
% If you have multiple appendixes use $\backslash${\tt{appendices}} then use $\backslash${\tt{section}} to start each appendix.
% You must declare a $\backslash${\tt{section}} before using any $\backslash${\tt{subsection}} or using $\backslash${\tt{label}} ($\backslash${\tt{appendices}} by itself
%  starts a section numbered zero.)}
